%% file: emnlp2023.tex
\pdfoutput=1

\documentclass[11pt]{article}
\usepackage{EMNLP2023}

\usepackage{times}
\usepackage{latexsym}
\usepackage[T1]{fontenc}
\usepackage[utf8]{inputenc}
\usepackage{microtype}
\usepackage{inconsolata}
\usepackage{booktabs}
\usepackage{multirow}
\usepackage{amsmath}
\usepackage{amssymb}
\usepackage{graphicx}
\usepackage{etoolbox}
\usepackage{url}
\usepackage{dblfloatfix}
\usepackage{placeins}
\usepackage[hypcap=false]{caption}
\usepackage[most]{tcolorbox}
\usepackage[table]{xcolor}
\usepackage{enumitem}
\makeatletter
\def\input@path{{./}{../}}
\makeatother
\graphicspath{{./}{../}{figures/}{../figures/}}
\definecolor{RiskHigh}{HTML}{C0392B}
\definecolor{RiskMed}{HTML}{D68910}
\definecolor{RiskLow}{HTML}{2E7D32}
\newcommand{\whigh}[1]{\textcolor{RiskHigh}{\textbf{#1}}}
\newcommand{\wmed}[1]{\textcolor{RiskMed}{\textbf{#1}}}
\newcommand{\wlow}[1]{\textcolor{RiskLow}{\textbf{#1}}}
\usepackage{fvextra}
\DeclareUnicodeCharacter{2013}{-}
\DeclareUnicodeCharacter{2014}{---}
\DeclareUnicodeCharacter{2019}{'}
\DeclareUnicodeCharacter{2192}{$\rightarrow$}
\DeclareUnicodeCharacter{2713}{\checkmark}

\title{Beyond Semantic Accuracy: Consequence-Aware Evaluation for Safety-Critical Language Understanding}

\author{
\normalfont
\textbf{Yujing Chang}\textsuperscript{1},
\textbf{Thinh Pham}\textsuperscript{2},
\textbf{Van-Phat Thai}\textsuperscript{1},
\textbf{Chunyao Ma}\textsuperscript{1}\\
\textbf{Yash Guleria}\textsuperscript{3},
\textbf{Pham Nhut Huy}\textsuperscript{1},
\textbf{Sameer Alam}\textsuperscript{1}\\
\textsuperscript{1}ATMRI, Nanyang Technological University (NTU), Singapore\\
\textsuperscript{2}Centre of AI Research, VinUniversity, Vietnam\\
\textsuperscript{3}School of Management, Indian Institute of Technology Mandi, India
}

\makeatletter
\let\@oldmaketitle\@maketitle
\renewcommand{\@maketitle}{%
  \@oldmaketitle
  \vspace{-8pt}
  \centering
  \makebox[\linewidth][c]{\includegraphics[width=1.08\linewidth, trim={0.0em 0em 0.5em 0.0em}, clip]{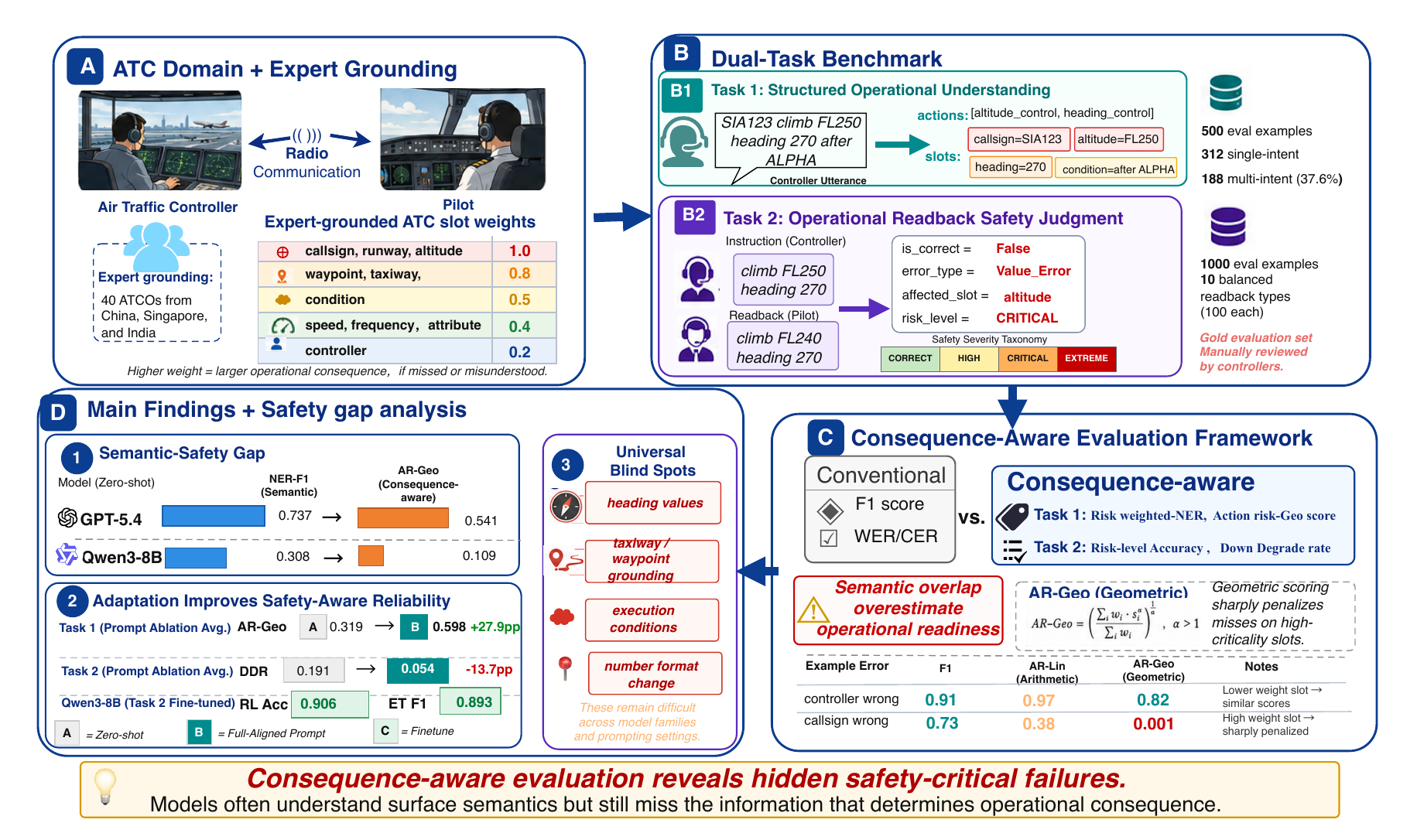}}
  \vspace{-1em}
  \captionof{figure}{Overview of our consequence-aware evaluation framework. An ATC utterance is mapped to an action schema with consequence-weighted slots; a structured understanding model (Task~1) and a readback safety judgment model (Task~2) are evaluated against that schema using nonlinear metrics that penalize safety-critical misses more severely than surface errors.}
  \label{fig:framework-overview}
  \vspace{17pt}
}
\makeatother
\begin{document}
\maketitle

\begin{abstract}
Can language models be trusted in safety-critical operations?
In such settings, strong performance on semantic metrics does not guarantee operational reliability: a misread altitude, a dropped execution condition, or a confused callsign may score well under standard F1 yet carry sharply asymmetric operational consequences.
We study this problem in air traffic control (ATC), where controller-pilot communication demands near-zero error tolerance, and use consequence-aware evaluation to test whether semantic scores misstate operational reliability.
The framework is instantiated in a controlled diagnostic ATC benchmark grounded in aviation standards and feedback from 40 air traffic controllers across three countries.
Evaluating 8 models, we uncover a systematic semantic-safety gap: conventional scores give substantially higher performance estimates than consequence-aware evaluation, even for models that appear reliable under standard metrics.
Risk-aware fine-tuning narrows but does not close this gap, showing that consequence-aware evaluation is a necessary complement to standard NLP metrics before any real safety-critical deployment claim.
\end{abstract}

\input{Tex/1Introduction}
\input{Tex/2RelatedWork}
\input{Tex/3Dataset}
\input{Tex/4Evaluation}
\input{Tex/5Experiment}

\input{Tex/6analysis}
\input{Tex/7Conclusion}

\clearpage
\bibliographystyle{acl_natbib}
\bibliography{custom,anthology}

\nolinenumbers
\appendix
\input{Tex/AAppendix}

\end{document}

%% file: Tex/1Introduction.tex
\section{Introduction}
\label{sec:intro}

Language models have rapidly moved from open-ended chat to practical assistance in writing, coding, search, and decision support~\cite{openai2024gpt4technicalreport,anthropic2024claude3modelcard,ouyang2022training}.
As these systems enter operational workflows, average semantic accuracy becomes an incomplete proxy for safety-critical reliability: a harmless formatting difference, a recoverable contextual miss, and a wrong control value may score similarly despite very different consequences.

Safety-critical language understanding is therefore a problem of consequence under operational risk, not just semantic similarity.
Standard evaluation for speech and task-oriented language understanding often relies on surface or semantic metrics such as word error rate (WER), character error rate (CER), intent accuracy, and slot-level F1~\cite{graves2006ctc,graves2014speech,hemphill1990atis,henderson2014second}.
These metrics typically treat errors uniformly across tokens, slots, or intents, creating a mismatch when action targets, control values, or execution conditions carry different operational consequences.

We study this problem in air traffic control (ATC), where controller-pilot coordination is voice-based and must be interpreted with high reliability~\cite{icao2016doc4444,connell1994pilotcontroller}.
ATC utterances are short, regulated, and operationally dense: confusing an altitude, substituting a callsign, or dropping a condition such as \textit{``after passing RIVER''} can contribute to loss of separation, runway incursions, or traffic conflicts, yet conventional metrics can obscure these differences~\cite{monan1983addressee,connell1994pilotcontroller}.

Our primary contribution is the empirical demonstration that conventional semantic evaluation can systematically misrepresent operational reliability in asymmetric-risk communication.
We make this gap measurable by mapping language-understanding errors to action structure, critical information units, and consequence-sensitive scores, with criticality definitions grounded in aviation standards and real-expert validation.
Our evaluation asks two questions: whether models recover the information whose failure would matter most in operation, and whether they can judge the safety significance of controlled readback variations.
Figure~\ref{fig:framework-overview} summarizes this pipeline.

The framework is intentionally simple: it exposes failures that standard semantic metrics leave under-specified while remaining adaptable to domains that can define actions, critical slots, and consequence levels.
Our results suggest that consequence-aware evaluation reveals a class of model failures that standard NLP metrics routinely underestimate.
We argue that safety-critical evaluation should explicitly account for operational consequence, and we release the data, annotation schemas, evaluation code, and prompts to support future work in this direction.\footnote{Artifacts are available at \url{https://github.com/EthanChangCC/beyond-semantic-accuracy}.}

\paragraph{Contributions.}
We make four concrete contributions:

\begin{itemize}[leftmargin=*, itemsep=0.3em, topsep=0.2em]
\item \textbf{Semantic-safety gap in operational language understanding.}
We show that conventional semantic metrics can substantially overestimate operational reliability in asymmetric-risk communication.
\item \textbf{Consequence-aware evaluation framework.}
We introduce an ATC evaluation framework that models action-critical intents, slots, execution conditions, and non-compensatory operational failures.
\item \textbf{Expert-grounded diagnostic ATC benchmark.}
We release a controlled dual-task ATC diagnostic benchmark grounded in aviation standards and expert-reviewed annotations that keep the diagnostic cases close to operational practice.
The benchmark tests both structured language understanding and real-operation error detection.
\item \textbf{Systematic model evaluation and adaptation study.}
Across prompting, fine-tuning, and model scales, we demonstrate substantial divergence between semantic correctness and operational safety.
\end{itemize}

%% file: Tex/2RelatedWork.tex
\section{Related Work}
\label{sec:related}

\paragraph{Language understanding and spoken language understanding.}
Task-oriented language understanding has traditionally been evaluated through intent accuracy, slot-level F1, and exact match~\cite{hemphill1990atis,henderson2014second}.
Much of the SLU literature therefore improves semantic prediction under noisy or complex input, including ASR correction and domain adaptation~\cite{mani2020asrerrorcorrectiondomain,mani-etal-2020-towards}, robustness to ASR-corrupted transcripts~\cite{chang2022contrastive,cheng2023c2aslu,cheng-etal-2023-ml,cheng-etal-2024-moe}, speech-text representation learning~\cite{chen2023salm,dong-etal-2023-cif}, multi-intent intent-slot modeling~\cite{qin-etal-2020-agif,Yin_Huang_Xu_2024,yin-etal-2025-eclm}, and LLM-based noisy slot filling or intent detection~\cite{sun-etal-2024-speech,he2023chatgptdetectintent}.
These methods improve semantic robustness, but their objectives remain slot accuracy, SLU-F1, or intent classification rather than operational consequence.

\paragraph{Risk-aware and safety-oriented evaluation.}
Standard accuracy can overestimate model reliability, motivating behavioral tests and severity-aware metrics such as CheckList and SEScore~\cite{ribeiro-etal-2020-beyond,xu-etal-2022-errors}.
Recent LLM work studies uncertainty estimation, safety prompting and datasets, risk awareness in agent interactions, and outcome-aware safety failures such as consequence-blindness~\cite{huang2025uncertainty,rottger2025safetyprompts,yuan-etal-2024-r,wu2025readscenescriptoutcomeaware}.
Safety evaluation has also expanded in operational domains such as medical task-oriented dialogue, radiology report generation, and clinical LLM benchmarking~\cite{saley-etal-2024-meditod,guan2025radiologyrisk,wang2026medicalbenchmark}.
These studies move beyond aggregate accuracy, but typically focus on general behavioral failures, harmful content, uncertainty, or domain-level safety labels; we quantify how structured semantic errors map to concrete operational risk.

\paragraph{ATC and aviation language understanding.}
ATC language processing has primarily focused on corpora, ASR, and domain-specific SLU.
Resources such as ATIS and ATCOSIM established spoken aviation datasets~\cite{hemphill1990atis,hofbauer-etal-2008-atcosim}, followed by ASR benchmarks and larger ATC communication corpora~\cite{zuluagagomez2020atcasr,zuluagagomez2023atco2,lin2021atcspeech}.
Recent work addresses accented ATC ASR, route inference from progressive taxi instructions, communication-error detection, and broader aviation-agent safety or rule-learning tasks~\cite{wee2025accentedatc,thai2025speechtoroute,SEMIHSADAK2026132241,wu2026pilotbench,wang2026world2rules}.
Prior work studies safety-oriented evaluation for ATC language understanding in an initial, smaller-scale setting with a simpler scoring formulation~\cite{chang2026safetyorientedevaluationlanguageunderstanding}.
We extend this direction with a broader consequence-aware diagnostic framework, combining controller-grounded validation, non-compensatory AR-Geo scoring, larger controlled datasets, dual structured-understanding and readback-safety tasks, and risk-aware adaptation experiments.

%% file: Tex/3Dataset.tex
\section{Dual-Task Dataset}
\label{sec:dataset}

We construct a controlled diagnostic ATC dataset grounded in aviation standards and validated by real air traffic controllers (details in Section~\ref{sec:framework} and Appendix~\ref{app:questionnaire}): Task~1 tests structured recovery of controller intent and slots, while Task~2 tests readback safety judgment under targeted perturbations.
The action and slot distribution—with altitude control, clearance, and heading control dominating and execution conditions present in over half of examples—reflects real operational patterns rather than uniform sampling; full statistics are in Appendix~\ref{app:dataset-details}.
Appendix~\ref{app:iaa} reports inter-annotator agreement for action labels, slot spans, readback risk levels, and error types; Appendix~\ref{app:case-validation} gives case-level validation examples.

\begin{table*}[t]
\centering
\scriptsize
\setlength{\tabcolsep}{2.2pt}
\begin{tabular}{lcccp{3.85cm}|llccc|llccc}
\toprule
\textbf{Action} & \textbf{w} & \textbf{N} & \textbf{\%} & \textbf{Scored slots} &
\textbf{Role} & \textbf{Slot} & \textbf{w} & \textbf{N} & \textbf{\%} &
\textbf{Role} & \textbf{Slot} & \textbf{w} & \textbf{N} & \textbf{\%} \\
\midrule
altitude\_control & \whigh{1.00} & 169 & 33.8 & callsign, altitude, condition & Target & callsign & \whigh{1.00} & 446 & 89.2 & Value & frequency & \wlow{0.40} & 54 & 10.8 \\
clearance & \whigh{1.00} & 161 & 32.2 & callsign, runway, waypoint, condition & Target & runway & \whigh{1.00} & 170 & 34.0 & Value & speed & \wlow{0.40} & 44 & 8.8 \\
runway\_instruction & \whigh{1.00} & 50 & 10.0 & callsign, runway, condition & Value & altitude & \whigh{1.00} & 164 & 32.8 & Context & relation & \wlow{0.40} & 92 & 18.4 \\
heading\_control & \wmed{0.60} & 141 & 28.2 & callsign, heading, condition & Value & heading & \wmed{0.80} & 99 & 19.8 & Context & attribute & \wlow{0.40} & 37 & 7.4 \\
taxiway\_instruction & \wmed{0.80} & 36 & 7.2 & callsign, taxiway, waypoint, runway & Target & waypoint & \wmed{0.80} & 91 & 18.2 & Context & traffic & \wlow{0.40} & 10 & 2.0 \\
speed\_control & \wmed{0.60} & 71 & 14.2 & callsign, speed, condition & Target & taxiway & \wmed{0.80} & 13 & 2.6 & Context & controller & \wlow{0.20} & 112 & 22.4 \\
contact & \wlow{0.30} & 59 & 11.8 & callsign, frequency, controller & Condition & condition & \wmed{0.50} & 264 & 52.8 & Other & O & \wlow{0.00} & 306 & 61.2 \\
sequence\_control & \wlow{0.40} & 46 & 9.2 & callsign, condition, traffic & & & & & & & & & & \\
\bottomrule
\end{tabular}
\caption{Task~1 action-risk schema and slot inventory.  expert-derived consequence weight; action  action instances, while slot  utterances containing the slot at least once. Counts are non-exclusive.}
\label{tab:task1-actions}
\label{tab:task1-slots}
\label{tab:slot-weights}
\end{table*}

\subsection{Task 1: Structured Operational Understanding}

\paragraph{Source data.}
We use the consolidated public ATC speech dataset from prior Speech-to-Route work~\cite{thai2025speechtoroute}, which covers U.S. and European controller-pilot communications.
After filtering incomplete or non-operational exchanges, we sample 500 structurally complete utterances for evaluation.

\paragraph{Gold evaluation annotations.}
Five air traffic controllers manually annotate the 500-utterance evaluation set with utterance-level \textit{actions} and chunk-level labels for operational spans.
Table~\ref{tab:task1-actions} reports the action inventory, slot inventory, coverage, and expert-derived weights.
For adaptation, we use separate training and validation splits following the same schema; these are used only for fine-tuning, not evaluation.
Additional details are in Appendix~\ref{app:dataset-details}.

\begin{table*}[t]
\centering
\scriptsize
\setlength{\tabcolsep}{3pt}
\begin{tabular}{lllp{9.5cm}}
\toprule
\textbf{Type} & \textbf{Risk} & \textbf{N} & \textbf{Description and Example} \\
\midrule
\textsc{value\_critical} & \textsc{critical} & 100 & Critical value wrong: ``climb FL250'' $\rightarrow$ ``climb FL350''. \\
\textsc{omission\_critical} & \textsc{critical} & 100 & Critical slot omitted: ``climb FL250 heading 270'' $\rightarrow$ ``heading 270''. \\
\textsc{target\_critical} & \textsc{critical} & 100 & Callsign substituted or confused: ``BAW123'' $\rightarrow$ ``BAW132''. \\
\textsc{value\_high} & \textsc{high} & 100 & Non-critical value wrong, such as speed or frequency. \\
\textsc{omission\_high} & \textsc{high} & 100 & Non-critical slot omitted, such as speed or frequency. \\
\textsc{constraint\_high} & \textsc{high} & 100 & Execution condition omitted: ``climb FL250 before ALPHA'' $\rightarrow$ ``climb FL250''. \\
\textsc{multi\_error} & \textsc{extreme} & 100 & Two or more critical slots wrong simultaneously. \\
\textsc{filler\_correct} & \textsc{correct} & 100 & Correct readback with filler words; operational meaning unchanged. \\
\textsc{number\_format} & \textsc{correct} & 100 & Non-standard number format; value remains semantically identical. \\
\textsc{correct} & \textsc{correct} & 100 & Exact correct readback. \\
\bottomrule
\end{tabular}
\caption{Task~2 readback safety taxonomy. Type is the controlled perturbation, Risk is gold severity, and $N$ is examples per type.}
\label{tab:task2-types}
\end{table*}

\subsection{Task 2: Operational Readback Safety Judgment}

\paragraph{Task design.}
Task~2 goes beyond slot extraction in a diagnostic setting: a model must decide whether a controlled pilot-readback variant preserves the safety-critical meaning of the original clearance.

\paragraph{Dataset construction.}
We derive Task~2 from Task~1 by constructing readbacks with controlled perturbations based on ICAO readback requirements, controller feedback, and common failure modes~\cite{icao2016doc4444}.
Critical errors target altitude, heading, runway, and callsign; high-risk errors target lower-criticality values or execution constraints; correct variants include harmless filler words and number formatting.
The evaluation set is intentionally balanced to ensure each error category is reliably assessed; it is a controlled diagnostic tool rather than a frequency-matched sample of live traffic, and it should not be used to estimate live readback-error frequencies or deployment readiness.
A separate 2853-example training set is used only for fine-tuning.

\paragraph{Error taxonomy and statistics.}
The Task~2 evaluation set contains 1000 examples, balanced across ten readback types and four risk levels: \textsc{correct}, \textsc{high}, \textsc{critical}, and \textsc{extreme}.

%% file: Tex/4Evaluation.tex
\section{Consequence-Aware Evaluation Framework}
\label{sec:framework}

We evaluate whether a model recovers the components whose failure would change the operational consequence of an ATC instruction or readback.
The framework has three components: weighted semantic matching, nonlinear completeness scoring, and ordinal downgrade penalties.

\subsection{Expert Grounding}

The criticality scheme is grounded in aviation procedure~\cite{icao2016doc4444} and ratings from 40 air traffic controllers across China, Singapore, and India.
Controllers rated the severity of missing or misinterpreting ATC information types and reviewed the Task~2 readback-error categories.
We normalize mean severity ratings within each question block, discretize them into the tiered Task~1 weights in Table~\ref{tab:task1-slots}, and report the questionnaire protocol, medians, IQRs, and sample variances in Appendix~\ref{app:questionnaire}.
The weights encode relative operational criticality rather than accident probability.
Appendix~\ref{app:sensitivity} reports sensitivity analyses over smoothing constants, slot weights, schema strictness, and aggregation choices; the main rankings remain stable.
To reduce circularity, we also compare AR-Geo against NER-F1, rNER-F1, and AR-Lin on independent controller acceptability ratings; AR-Geo has the highest alignment (Pearson $r=0.68$, Spearman $\rho=0.65$; Table~\ref{tab:human-alignment}).

\subsection{Task 1: Structured Operational Understanding}

Because utterances may contain multiple actions, \textit{Action-Exact} requires the full predicted action set to exactly match the gold set.

\paragraph{Entity scoring.}
We report token-level NER-F1 over all slot types with equal weight.
NER-Lin uses the same exact span-and-type matching rule, weighting each matched gold or predicted span by the slot weight in Table~\ref{tab:task1-slots}; full matching details are in Appendix~\ref{app:metric-details}.
NER-Geo is a recall-oriented geometric score over gold spans:
\begin{equation}
\mathrm{NER\text{-}Geo} =
\exp\!\left(
\frac{\sum_{s\in\mathcal{G}} w(s)\log(\epsilon+m_s)}
{\sum_{s\in\mathcal{G}} w(s)}
\right),
\label{eq:ner-geo}
\end{equation}
where $m_s=1$ if gold span $s$ is exactly matched and $0$ otherwise.
We set $\epsilon=10^{-5}$ to avoid an undefined logarithm.

\paragraph{Action scoring.}
Entity scores do not bind slots to actions, so we compute an action-conditioned score over slots required by each gold action schema.
For each action instance $a$, let $S_a$ be the set of required or observed gold slots associated with that action, and let $m_i\in\{0,1\}$ indicate whether slot $i\in S_a$ is correctly recovered.
If the action type itself is not predicted, the action instance receives score zero.
For multi-intent utterances, we score each gold action instance and average using the action weights in Table~\ref{tab:task1-actions}.

The linear variant, AR-Lin, is:
\begin{equation}
\mathrm{AR\text{-}Lin}(a) =
\frac{\sum_{i\in S_a} w_i m_i}{\sum_{i\in S_a} w_i}.
\label{eq:ar-lin}
\end{equation}
AR-Lin is a weighted baseline, but lower-risk matches can still compensate for a high-risk miss. We therefore use AR-Geo as a non-compensatory diagnostic score: a missed high-consequence field should not be averaged away by many lower-risk matches. Inspired by BLEU's geometric aggregation~\cite{papineni2002bleu} and reliability analysis~\cite{nrc1975reactor}, we use:

\begin{equation}
\mathrm{AR\text{-}Geo}(a) =
\exp\!\left(
\frac{\sum_{i\in S_a} w_i \log(\epsilon + m_i)}
{\sum_{i\in S_a} w_i}
\right),
\label{eq:geo}
\end{equation}
Utterance-level AR-Geo is the weighted average over gold action instances.
AR-Geo penalizes missing required components without introducing a tunable severity coefficient.
Table~\ref{tab:motivating} gives an example and shows that AR-Geo has the strongest empirical correlation with controller ratings among the tested alternatives. The questionnaire-based paired-preference study in Appendix~\ref{app:controller-preference} provides additional support: AR-Geo matches controller preferences more often than AR-Lin when surface overlap and operational safety diverge (0.76 vs.\ 0.64).

\subsection{Task 2: Readback Safety Judgment}

Task~2 asks whether a pilot readback preserves operational safety.
For each instruction-readback pair, the model predicts correctness, error type, affected slot, and risk level.
The risk levels are ordered as:
\[
\textsc{correct} < \textsc{high} < \textsc{critical} < \textsc{extreme}.
\]

We report \textit{isCorrect accuracy}, \textit{error-type macro-F1} (ET~F1), and \textit{risk-level accuracy} (RL~Acc) for binary correctness, error mechanism, and severity assignment.

\paragraph{Directional safety calibration.}
Risk-level errors are directional: under-estimating risk can hide a dangerous readback error, so we report two lower-is-better downgrade metrics.
Let $r_i$ and $\hat{r}_i$ denote the ordinal gold and predicted risk levels for example $i$, and let $R_{\max}$ be the maximum possible ordinal gap.
Dangerous Downgrade Rate (DDR) measures how often the model predicts a lower risk level than the gold label:
\begin{equation}
\mathrm{DDR} =
\frac{|\{i:\hat{r}_i < r_i\}|}{N}.
\label{eq:ddr}
\end{equation}
Weighted Downgrade Severity (WDS) measures normalized downgrade mass over all examples, assigning zero contribution to non-downgraded cases:
\begin{equation}
\mathrm{WDS} =
\frac{1}{N}\sum_{i:\hat{r}_i < r_i}
\frac{r_i-\hat{r}_i}{R_{\max}}.
\label{eq:wds}
\end{equation}
Thus, predicting \textsc{correct} for an \textsc{extreme} error is worse than predicting \textsc{high} for a \textsc{critical} error.

\begin{table}[t]
\centering
\scriptsize
\setlength{\tabcolsep}{3pt}
\begin{tabular}{p{4.2cm}cc}
\toprule
\textbf{Prediction error} & \textbf{AR-Lin} & \textbf{AR-Geo} \\
\midrule
Controller wrong: ``contact singapore ground'' $\to$ ``singapore tower'' & 0.92 & 0.39 \\
Altitude wrong: ``descend flight level 240'' $\to$ ``descend flight level 250'' & 0.68 & 0.02 \\
\bottomrule
\end{tabular}

\vspace{0.7em}
\begin{tabular}{lccc}
\toprule
\textbf{Metric} & \textbf{Pearson $r$} & \textbf{Spearman $\rho$} & \textbf{$p$-value} \\
\midrule
NER-F1 & 0.44 & 0.42 & $<0.001$ \\
rNER-F1 & 0.53 & 0.51 & $<0.001$ \\
AR-Lin & 0.59 & 0.57 & $<0.001$ \\
AR-Geo & \textbf{0.68} & \textbf{0.65} & $<0.001$ \\
\bottomrule
\end{tabular}
\caption{Nonlinear scoring validation. Top: AR-Lin vs.\ AR-Geo; bottom: correlation with controller-rated acceptability on 200 Task~1 outputs.}
\label{tab:motivating}
\label{tab:human-alignment}
\label{tab:human_alignment}
\end{table}

%% file: Tex/5Experiment.tex
\section{Experimental Setup}
\label{sec:exp}

Task~1 evaluates eight commercial and open-weight models for broad zero-shot coverage.
Controlled prompting and Task~2 use a four-model subset spanning API and open-weight settings.
Unless marked as fine-tuned, models are used through their standard inference interfaces with greedy decoding (temperature~0).
Reported model identifiers are the provider or local model strings recorded in the inference outputs; API snapshot names and execution dates are listed in Appendix~\ref{app:model-versioning}.
Task~2 outputs \texttt{is\_correct}, \texttt{error\_type}, \texttt{risk\_level}, \texttt{affected\_slot}, and \texttt{explanation}; invalid or unparseable responses receive zero for the relevant metrics.
All conditions use the 500-example Task~1 held-out set and the 1000-example Task~2 readback evaluation set.

\paragraph{Prompting conditions.}
\label{subsec:prompt}
We compare four prompting conditions: Zero-shot (A) provides only the schema and field definitions; Knowledge (C) adds ATC operational rules; Few-shot (D) adds annotated examples; and Full-Aligned (B) combines rules and examples.
For Task~2, we also evaluate Few-shot+CoT~\cite{wei2022chain} to test whether explicit reasoning changes risk calibration.
Appendix~\ref{app:prompts} lists the full prompt templates and output contracts.

\paragraph{Fine-tuning baseline.}
\label{subsec:ft}

To estimate domain-adaptation headroom, we fine-tune Qwen3-8B~\cite{qwen3} with LoRA~\cite{hu2022lora}.
For Task~1, we compare standard cross-entropy (CE) with Risk-Loss, a consequence-weighted variant that up-weights critical-slot tokens such as altitude, heading, runway, and callsign.
For Task~2, we train with an analogous weighted cross-entropy objective that prioritizes risk level, error type, and higher-severity samples.
For both tasks, the risk-aware objective replaces mean cross-entropy with weighted cross-entropy:
\begin{equation}
\mathcal{L}_{\mathrm{risk}} =
\frac{\sum_i \alpha_i\,\mathrm{CE}(z_i,y_i)\mathbf{1}[y_i\neq -100]}
{\sum_i \alpha_i\mathbf{1}[y_i\neq -100]},
\label{eq:risk-loss}
\end{equation}
where $\alpha_i$ is a token-level weight derived from slot criticality or readback-risk severity.
Appendix~\ref{app:finetune} gives the LoRA configuration, training hyperparameters, and exact weighting rules.

\paragraph{Evaluation protocol.}

We follow the metric definitions in Section~\ref{sec:framework}.
For Task~1, AR-Geo is the primary consequence-aware structured-understanding metric.
For Task~2, we emphasize error-type F1, risk-level accuracy, Dangerous Downgrade Rate (DDR), and Weighted Downgrade Severity (WDS).
DDR and WDS measure dangerous under-estimation and are lower-is-better.

\paragraph{Research questions.}
We ask how large the semantic-safety gap is (RQ1), which slots and readback types drive it (RQ2), which adaptation strategy helps most (RQ3), and whether Task~1 structured understanding transfers to Task~2 safety judgment (RQ4).

%% file: Tex/6analysis.tex
\section{Results and Analysis}
\label{sec:analysis}

\subsection{Task 1: Structured Understanding (RQ1, RQ2, RQ4)}
\label{subsec:task1-results}

\begin{table*}[!t]
\centering
\renewcommand{\arraystretch}{1.05}
\setlength{\tabcolsep}{2.0pt}
\scriptsize
\begin{tabular}{llcccccccc}
\hline
\textbf{Model} & \textbf{Setting} & \textbf{NER-F1} & \textbf{NER-Lin} & \textbf{NER-Geo} & \textbf{ActEx} & \textbf{AR-Lin} & \textbf{AR-Geo}$^\star$ & \textbf{Strict} & \textbf{Lat.(s)} \\
\hline
\multicolumn{10}{l}{\textit{Zero-shot frontier comparison}} \\
gpt-5.4          & ZS & \textbf{0.737} & \textbf{0.768} & \textbf{0.443} & \textbf{0.494} & \textbf{0.774} & \textbf{0.541} & \textbf{0.442} & 2.3 \\
gpt-5.1          & ZS & 0.720 & 0.752 & 0.415 & 0.474 & 0.722 & 0.451 & 0.360 & 2.3 \\
DeepSeek-V4-Flash    & ZS & 0.667 & 0.699 & 0.404 & 0.448 & 0.746 & 0.487 & 0.414 & 2.5 \\
claude-haiku-4.5 & ZS & 0.634 & 0.667 & 0.347 & 0.468 & 0.719 & 0.431 & 0.362 & $-^\dagger$ \\
gpt-4o-mini      & ZS & 0.647 & 0.690 & 0.331 & 0.354 & 0.667 & 0.338 & 0.258 & 2.5 \\
qwen-plus        & ZS & 0.644 & 0.676 & 0.330 & 0.454 & 0.639 & 0.343 & 0.278 & 3.5 \\
qwen3-14b        & ZS & 0.546 & 0.573 & 0.234 & 0.392 & 0.561 & 0.255 & 0.196 & 3.4 \\
qwen3-8b         & ZS & 0.308 & 0.330 & 0.076 & 0.432 & 0.290 & 0.109 & 0.080 & 4.4 \\
\hline
\multicolumn{10}{l}{\textit{Prompt-enhanced comparison (Full-Aligned)}} \\
DeepSeek-V4-Flash    & \quad - \textit{After Full-Aligned prompt} & \textbf{0.773} & 0.798 & \textbf{0.555} & \textbf{0.612} & \textbf{0.836} & \textbf{0.707} & \textbf{0.614} & 2.3 \\
gpt-4o-mini      & \quad - \textit{After Full-Aligned prompt} & 0.751 & 0.784 & 0.490 & 0.576 & 0.775 & 0.589 & 0.512 & 2.9 \\
qwen-plus        & \quad - \textit{After Full-Aligned prompt} & \textbf{0.773} & \textbf{0.800} & 0.528 & 0.582 & 0.783 & 0.609 & 0.530 & 4.0 \\
qwen3-8b         & \quad - \textit{After Full-Aligned prompt} & 0.684 & 0.708 & 0.429 & 0.548 & 0.701 & 0.489 & 0.372 & 3.2 \\
\hline
\multicolumn{10}{l}{\textit{Fine-tuning comparison with matched zero-shot baselines}} \\
qwen3-4b         & \quad - \textit{ZS} & 0.213 & 0.232 & 0.037 & 0.316 & 0.118 & 0.041 & 0.024 & -- \\
                 & \quad - \textit{FT (Risk-LoRA)} & 0.762 & 0.783 & 0.554 & 0.606 & 0.777 & 0.648 & 0.538 & -- \\
qwen3-8b         & \quad - \textit{ZS} & 0.308 & 0.330 & 0.076 & 0.432 & 0.290 & 0.109 & 0.080 & 4.4 \\
                 & \quad - \textit{FT (CE loss)} & 0.700 & 0.731 & 0.424 & 0.464 & 0.746 & 0.515 & 0.396 & 6.9 \\
                 & \quad - \textit{FT (Risk-LoRA)} & \textbf{0.772} & \textbf{0.792} & \textbf{0.570} & \textbf{0.638} & 0.805 & 0.686 & 0.584 & 5.5 \\
Llama-3.1-8B     & \quad - \textit{ZS} & 0.223 & 0.246 & 0.048 & 0.378 & 0.197 & 0.057 & 0.032 & -- \\
                 & \quad - \textit{FT (Risk-LoRA)} & 0.767 & 0.788 & 0.559 & 0.620 & \textbf{0.813} & \textbf{0.697} & \textbf{0.600} & -- \\
\hline
\end{tabular}
\caption{Task~1 main results (N=500). ActEx=action-set exact match; Strict=all gold actions and action-conditioned slots recovered; Lat.=median seconds/sample. ZS=Zero-shot, FA=Full-Aligned, FT=LoRA. $^\dagger$Not recorded.}
\label{tab:task1-main}
\end{table*}

\begin{table}[!b]
\centering
\scriptsize
\setlength{\tabcolsep}{2.5pt}
\begin{tabular}{lccccc}
\toprule
\textbf{Cond.} & \textbf{AR-Geo}$^\star$ & \textbf{ActEx} & \textbf{Head} & \textbf{Cond} & \textbf{CritV} \\
\midrule
A (ZS)    & 0.319 & 0.422 & 0.209 & 0.188 & 0.382 \\
C (Know.) & 0.421 & 0.422 & 0.311 & 0.397 & 0.544 \\
D (Few)   & \textbf{0.638} & 0.554 & 0.740 & \textbf{0.591} & 0.826 \\
B (Full)  & 0.598 & \textbf{0.580} & \textbf{0.769} & 0.538 & \textbf{0.830} \\
\midrule
A$\to$B   & +27.9pp & +15.8pp & +56.0pp & +35.0pp & +44.8pp \\
\bottomrule
\end{tabular}
\caption{Task~1 prompt ablations averaged over four models. Head, Cond, and CritV are strict exact-span recalls for key slot types.}
\label{tab:task1-ablation}
\label{tab:task1-role-ablation}
\end{table}

\begin{figure*}[!t]
\centering
\includegraphics[width=\linewidth]{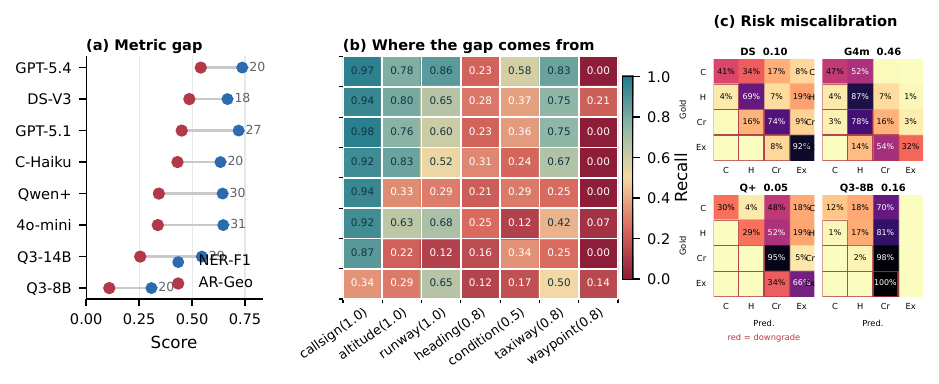}
\caption{Evaluation overview. (a) NER-F1 exceeds AR-Geo. (b) Strict boundary failures and value disappearance contribute to the gap, especially around heading-related spans and conditions. (c) Task~2 zero-shot risk calibration; C/H/Cr/Ex denote \textsc{correct}, \textsc{high}, \textsc{critical}, and \textsc{extreme}, and red outlines mark dangerous downgrades.}
\label{fig:task1-gap}
\label{fig:task2-confusion}
\end{figure*}

\begin{table*}[!t]
\centering
\renewcommand{\arraystretch}{1.0}
\setlength{\tabcolsep}{2.0pt}
\scriptsize
\begin{tabular*}{0.90\textwidth}{@{\extracolsep{\fill}}llcccccc@{}}
\hline
\textbf{Model} & \textbf{Setting} & \textbf{isCorr} & \textbf{ET F1} & \textbf{RL Acc} & \textbf{DDR}$\downarrow$ & \textbf{WDS}$\downarrow$ & \textbf{Lat.(s)} \\
\hline
\multicolumn{8}{l}{\textit{Zero-shot comparison}} \\
DeepSeek-V4-Flash & ZS & 0.774 & \textbf{0.563} & \textbf{0.646} & 0.100 & 0.034 & 1.5 \\
gpt-4o-mini   & ZS & \textbf{0.794} & 0.533 & 0.485 & 0.459 & 0.163 & 1.7 \\
qwen-plus     & ZS & 0.778 & 0.461 & 0.530 & 0.049 & 0.016 & 2.5 \\
qwen3-4b      & ZS & 0.714 & 0.170 & 0.330 & 0.240 & 0.090 & -- \\
qwen3-8b      & ZS & 0.720 & 0.268 & 0.382 & 0.157 & 0.052 & 1.7 \\
Llama-3.1-8B  & ZS & 0.706 & 0.286 & 0.244 & \textbf{0.031} & \textbf{0.011} & -- \\
\hline
\multicolumn{8}{l}{\textit{Prompt-enhanced comparison (Full-Aligned)}} \\
DeepSeek-V4-Flash & FA & \textbf{0.896} & \textbf{0.840} & \textbf{0.878} & 0.057 & 0.022 & $8.2^\dagger$ \\
gpt-4o-mini   & FA & 0.802 & 0.466 & 0.540 & 0.049 & 0.016 & 2.0 \\
qwen-plus     & FA & 0.812 & 0.565 & 0.640 & \textbf{0.023} & \textbf{0.009} & 2.6 \\
qwen3-8b      & FA & 0.758 & 0.360 & 0.526 & 0.089 & 0.030 & 2.1 \\
\hline
\multicolumn{8}{l}{\textit{Fine-tuning comparison}} \\
qwen3-4b      & FT & \textbf{0.946} & 0.914 & 0.930 & \textbf{0.060} & \textbf{0.020} & -- \\
qwen3-8b      & FT & 0.930 & 0.893 & 0.906 & 0.083 & 0.029 & 1.2 \\
Llama-3.1-8B  & FT & 0.944 & \textbf{0.931} & \textbf{0.936} & 0.080 & 0.027 & -- \\
\hline
\end{tabular*}
\caption{Task~2 main results (N=1000). isCorr=readback-correctness accuracy; ET F1=error-type macro-F1; RL Acc=risk-level accuracy; DDR and WDS are lower-is-better downgrade metrics; Lat.=median seconds/sample. $^\dagger$Extended thinking on a subset. ZS=Zero-shot, FA=Full-Aligned, FT=Risk-LoRA.}
\label{tab:task2-main}
\end{table*}

\paragraph{Semantic-safety gap.}
Table~\ref{tab:task1-main} and Figure~\ref{fig:task1-gap}a show a consistent semantic-safety gap: models recover many surface spans while missing components that make an instruction executable.
For example, gpt-5.4 reaches 0.737 NER-F1 under zero-shot prompting, while AR-Geo diagnoses a different failure mode: some recovered spans no longer make the instruction operationally complete. qwen3-8b shows the same diagnostic disagreement more sharply (0.308 NER-F1 vs.\ 0.109 AR-Geo).
Lower-risk matches therefore cannot compensate for missing high-consequence components; Appendix~\ref{app:argeo-collapse} and Table~\ref{tab:argeo-failure-examples} provide contrastive utterance-level examples showing that low-weight misses remain relatively tolerated, while missing a callsign or shared condition collapses AR-Geo despite substantial NER-F1.

\paragraph{What drives the gap.}
The gap is not evenly distributed: models are reliable on addressees and simple surface values but weaker on exact value boundaries, execution conditions, and route elements (Figure~\ref{fig:task1-gap}b).
For heading-related spans, we distinguish strict boundary failures from value disappearance: many outputs keep the numeric heading inside a neighboring action chunk, while more serious cases lose the value entirely (Appendix~\ref{app:task1-error-analysis}).
The prompt ablation confirms the pattern: moving from zero-shot to Full-Aligned prompting raises strict heading recall by 56.0pp and condition recall by 35.0pp, showing that models often know the broad action type while losing the value or condition that determines \textit{how} or \textit{when} it should be executed.

\paragraph{Human alignment.}
Controller ratings support the scoring design: consequence-aware scores track human judgments of operational acceptability more closely than standard semantic metrics (AR-Geo Pearson $r=0.68$ vs.\ NER-F1 $r=0.44$; Table~\ref{tab:human-alignment}), and rankings remain stable under weight and aggregation checks (Appendix~\ref{app:sensitivity}).
A paired controller-preference study shows the same pattern when controllers choose between outputs that trade surface overlap against safety-critical information (Appendix~\ref{app:controller-preference}).

\paragraph{Adaptation effects.}
Two patterns emerge from the new adaptation results. First, consequence-aware supervision substantially narrows the gap between open-weight models and stronger API systems. The Risk-LoRA variants of qwen3-4b, qwen3-8b, and Llama-3.1-8B reach 0.648--0.697 AR-Geo, exceeding all zero-shot frontier results in Table~\ref{tab:task1-main} and approaching the strongest Full-Aligned API score (0.707). This suggests that many Task~1 failures are not fixed properties of model scale; they reflect a mismatch between standard training objectives and the consequence structure of ATC commands.
Second, the objective matters after the model has learned the output format. On the same qwen3-8b backbone, Risk-LoRA improves over CE fine-tuning in strict exact recovery (0.584 vs.\ 0.396) and AR-Geo (0.686 vs.\ 0.515), while also raising NER-F1 (0.772 vs.\ 0.700). Consequence weighting therefore does not merely sacrifice semantic overlap for safety: it improves surface recovery while reallocating capacity toward high-criticality slots that determine operational completeness. Prompting remains useful but less reliable: Full-Aligned prompting helps action exact match and critical-value recall, whereas few-shot examples produce the highest average AR-Geo in the ablation; adding broad knowledge can introduce redundant reasoning and slightly reduce boundary-sensitive performance (Tables~\ref{tab:task1-main} and~\ref{tab:task1-ablation}). The persistent errors on waypoints and boundary-sensitive route structure show where adaptation still lacks enough operational coverage.

\subsection{Task 2: Readback Safety Judgment (RQ1--RQ4)}
\label{subsec:task2-results}

\paragraph{Risk miscalibration.}
Table~\ref{tab:task2-main} and Figure~\ref{fig:task2-confusion}c show that controlled readback safety judgment is not simply error detection.
Zero-shot models often identify the error mechanism, but risk calibration is unstable: gpt-4o-mini has the largest downgrade rate (DDR=0.459), while qwen-plus rarely downgrades (DDR=0.049) but still has limited risk-level accuracy (0.530), indicating a more conservative calibration pattern.

\paragraph{Error-type structure.}
The hardest cases echo Task~1: models are more reliable on explicit value or target substitutions than on execution constraints and harmless surface variation.
Per-error-type results in Appendix Table~\ref{tab:task2-detail} confirm that condition-related readback errors are consistently weaker than explicit value and target substitutions.
Readback judgment requires both sensitivity to dangerous omissions and invariance to equivalent paraphrases.

\paragraph{Adaptation effects.}
Task~2 separates error recognition from risk calibration. Explicit risk definitions improve prompting, but mainly test taxonomy-following risk assignment after error-type identification rather than unconstrained operational risk estimation (Table~\ref{tab:task2-main}; prompt ablations in Appendix Table~\ref{tab:task2-ablation}). DeepSeek-V4-Flash illustrates the limit: its Full-Aligned run is strong on isCorrect, error-type F1, and risk-level accuracy (0.896/0.840/0.878), yet still leaves nonzero downgrade mass.
Risk-aware fine-tuning is most effective on this calibration layer. Across open-weight backbones, risk-level accuracy rises from 0.244--0.382 in zero-shot use to 0.906--0.936 after fine-tuning; qwen3-4b leads isCorrect accuracy (0.946), while Llama-3.1-8B leads error-type F1 and risk-level accuracy (0.931/0.936). The remaining DDR/WDS and condition errors show that models can learn the taxonomy but still tend to treat execution conditions as modifiers rather than operational preconditions.

\subsection{Discussion}
\label{subsec:discussion}

\paragraph{Does Task~1 transfer to Task~2?}
Structured understanding helps but is not sufficient.
Across the four shared zero-shot models, Task~1 AR-Geo and Task~2 RL Acc are positively associated, yet similar extraction quality can hide very different dangerous-downgrade rates: gpt-4o-mini and qwen-plus reach nearly identical zero-shot AR-Geo (0.338 vs.\ 0.343) yet their DDR differs ninefold (0.459 vs.\ 0.049)---a gap invisible to extraction metrics alone.
Figure~\ref{fig:task1-task2-linkage} visualizes this split, showing that risk calibration is an additional capability beyond Task~1 extraction.

\begin{center}
\includegraphics[width=0.95\columnwidth]{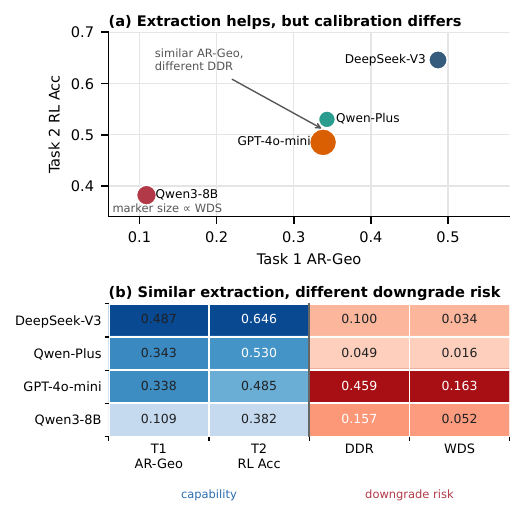}
\captionof{figure}{Cross-task diagnostic. Similar Task~1 AR-Geo can hide different Task~2 risk calibration and downgrade rates.}
\label{fig:task1-task2-linkage}
\end{center}

\paragraph{Implications.}
Execution conditions remain the clearest cross-task failure: ``climb to FL350'' and ``climb to FL350 after RIVER'' share most tokens but authorize different actions at different times.
Standard metrics are necessary but insufficient: surface accuracy can be high while operationally consequential information is lost, and closing that gap requires consequence-aware objectives, not just better prompts or larger models.

%% file: Tex/7Conclusion.tex
\section{Conclusion}
\label{sec:conclusion}

We empirically demonstrated a semantic-safety gap in safety-critical language understanding: standard semantic evaluation can systematically overestimate reliability when errors carry asymmetric consequences.
The consequence-aware framework and controlled ATC benchmark make this gap measurable across structured instruction understanding and readback safety judgment.

Prompting and risk-aware fine-tuning reduce but do not close this gap.
The contribution is a measurement and diagnosis framework, not a deployment-ready ATC system; the same recipe can transfer to domains where experts define actions, critical units, and severity levels (Appendix~\ref{app:transfer}).

\section*{Limitations}

ATC is a realistic high-stakes testbed, but the slot weights and Task~2 taxonomy are ATC-specific; adapting them requires new expert input and may change quantitative results.
Task~2 readbacks use structured perturbations, which support controlled diagnosis but do not fully capture live pilot readback variation; the Task~2 claims should therefore be read as diagnostic rather than frequency-matched operational prevalence estimates or evidence of deployment readiness.

Because of budget constraints, we do not exhaustively evaluate current commercial models or broad fine-tuning recipes.
Our fine-tuning study uses one 8B open-weight backbone, so model scale and data effects remain open.
Most importantly, improved scores are still far from deployment-grade ATC reliability.
Future work should pair stronger models with multi-layer safety mechanisms such as abstention, independent cross-checking, human review, and runtime monitoring.

\section*{Ethics Statement}

The ATC data is derived from publicly available or licensed corpora and contains no personal information about controllers or pilots.
The controller questionnaire was conducted on a voluntary basis for expert validation of evaluation criteria. Responses were anonymized before analysis, and no personally identifying information about respondents is released.
We identify safety-critical failure modes; none of the evaluated models should be deployed in actual ATC operations without additional validation.
Benchmark data, evaluation code, prompts, and fine-tuning artifacts are available in the public repository noted in the introduction. AI~assistance~was~used only for language polishing and grammar refinement. 

\section*{Acknowledgments}
This research is supported by the National Research Foundation, Singapore, and the Civil Aviation Authority of Singapore, under the Aviation Transformation Programme. Any opinions, findings and conclusions or recommendations expressed in this material are those of the author(s) and do not reflect the views of National Research Foundation, Singapore and the Civil Aviation Authority of Singapore.

%% file: Tex/AAppendix.tex
\section*{Appendix Overview}
\label{app:overview}

The appendix provides additional details supporting the consequence-aware evaluation framework, dataset design, prompting protocol, and model adaptation experiments.
It is organized as follows:
\begin{itemize}
    \item \textbf{Appendix~\ref{app:model-versioning}} reports the model identifiers and inference dates recorded from the output files and inference scripts.

    \item \textbf{Appendix~\ref{app:sensitivity}} reports sensitivity analyses for the consequence-aware scoring functions, including the geometric smoothing constant (Appendix~\ref{app:geo-eps}), slot-weight perturbations (Appendix~\ref{app:weight-sensitivity}), required-slot strictness (Appendix~\ref{app:schema-sensitivity}), and the exponential risk penalty as an alternative nonlinear score (Appendix~\ref{app:erp}).

    \item \textbf{Appendix~\ref{app:dataset-details}} describes the dataset and held-out evaluation sets, including the annotation protocol (Appendix~\ref{app:annotation-protocol}), inter-annotator agreement (Appendix~\ref{app:iaa}), label inventory and ambiguous cases (Appendix~\ref{app:label-inventory}), Task~1 operational-understanding statistics, Task~2 readback-verification statistics, and case-level validation examples (Appendix~\ref{app:case-validation}).

    \item \textbf{Appendix~\ref{app:metric-details}} gives the complete metric definitions and matching rules for NER-Lin, NER-Geo, AR-Lin, AR-Geo, Strict, missing actions, duplicate slots, bootstrap confidence intervals, and a worked example showing why nonlinear scoring is needed.

    \item \textbf{Appendix~\ref{app:task1-error-analysis}} provides additional Task~1 error analysis, especially the distinction between strict span failures and operational-value recovery for heading instructions.

    \item \textbf{Appendix~\ref{app:task2-taxonomy}} details the Task~2 error taxonomy, affected-slot rules, and examples for each readback category.

    \item \textbf{Appendix~\ref{app:finetune}} presents fine-tuning details, including SFT data format, Task~1 and Task~2 LoRA configurations, a comparison between CE-loss and risk-aware-loss Task~1 fine-tuning (AR-Geo 0.515 vs.\ 0.686), and the token-level risk-aware loss formulation.

    \item \textbf{Appendix~\ref{app:questionnaire}} describes the expert questionnaire instrument used to validate the evaluation criteria, including controller background statistics, the survey structure, and aggregation protocol for slot weights, action criticality, readback taxonomy, nonlinear metric preferences, and the paired controller-preference study (Appendix~\ref{app:controller-preference}).

    \item \textbf{Appendix~\ref{app:task2-detail}} provides additional Task~2 ablations and per-error-type results, highlighting which readback error categories remain difficult.

    \item \textbf{Appendix~\ref{app:transfer}} discusses how the consequence-aware evaluation framework can transfer beyond ATC by replacing the domain action inventory, critical slot mapping, and error taxonomy while keeping the scoring principle unchanged.

    \item \textbf{Appendix~\ref{app:prompts}} documents the prompting and inference protocol for the zero-shot, knowledge, few-shot, CoT, and full-aligned ablations.
\end{itemize}

\section{Model Identifiers and Inference Dates}
\label{app:model-versioning}

The API outputs record the requested model string and latency, but not a provider-side immutable snapshot hash.
We therefore report the exact model identifiers used by the inference scripts and the result-file modification dates as the execution record.
Provider-hosted aliases may change over time, so these results should be treated as reproducible with respect to the released prompts, data, and recorded identifiers, but not as guaranteed immutable API snapshots.

\begin{table*}[htbp]
\centering
\scriptsize
\setlength{\tabcolsep}{3pt}
\begin{tabular}{llll}
\toprule
\textbf{Task/setting} & \textbf{Reported model identifier} & \textbf{Inference interface} & \textbf{Recorded date} \\
\midrule
Task~1 zero-shot & gpt-5.4 & OpenAI-compatible API & 2026-04-22 \\
Task~1 zero-shot & gpt-5.1 & OpenAI-compatible API & 2026-04-22 \\
Task~1 zero-shot & gpt-4o-mini & OpenAI API & 2026-04-22 \\
Task~1 zero-shot & DeepSeek-V4-Flash & DeepSeek API & 2026-04-22 \\
Task~1 zero-shot & claude-haiku-4-5-20251001 & Anthropic API & 2026-04-23 \\
Task~1 zero-shot & qwen-plus & DashScope compatible API & 2026-04-22 \\
Task~1 zero-shot & qwen3-14b & DashScope compatible API & 2026-04-22 \\
Task~1 zero-shot & qwen3-8b & DashScope compatible API & 2026-04-22 \\
Task~1 Full-Aligned & DeepSeek-V4-Flash, gpt-4o-mini, qwen-plus & Provider APIs & 2026-05-01 \\
Task~1 Full-Aligned & qwen3-8b & DashScope compatible API & 2026-05-02 \\
Task~1 fine-tuned & qwen3-8b-atc-lora-fast & Local Qwen3-8B LoRA & 2026-05-22 \\
Task~1 fine-tuned & qwen3-8b-atc-task1-risk-lora-v3 & Local Qwen3-8B LoRA & 2026-05-25 \\
Task~2 zero-shot & DeepSeek-V4-Flash, gpt-4o-mini, qwen-plus & Provider APIs & 2026-04-28 \\
Task~2 zero-shot & qwen3-8b & DashScope compatible API & 2026-05-02 \\
Task~2 Full-Aligned & gpt-4o-mini, qwen-plus, qwen3-8b & Provider APIs & 2026-05-02 \\
Task~2 Full-Aligned & DeepSeek-V4-Flash & DeepSeek API & 2026-05-04 \\
Task~2 fine-tuned & qwen3-8b-task2-risk-lora & Local Qwen3-8B LoRA & 2026-05-12 \\
\bottomrule
\end{tabular}
\caption[Model identifiers and inference dates.]{Model identifiers and inference dates used in the reported experiments. Dates are taken from the corresponding output files; where the provider did not expose an immutable snapshot in the output, we report only the requested model identifier. Older DeepSeek output files used the legacy API alias \texttt{deepseek-chat}; we report these runs uniformly as DeepSeek-V4-Flash in all result tables.}
\label{tab:model-versioning}
\end{table*}

\section{Sensitivity Analysis for Consequence-Aware Scoring}
\label{app:sensitivity}

The central methodological choice in this paper is to evaluate safety-critical language understanding with a nonlinear consequence-aware score.
This appendix reports additional sensitivity checks showing that the main conclusions are not artifacts of a particular numerical constant, slot-weight vector, or nonlinear family.
We evaluate four perturbations: the geometric smoothing constant $\epsilon$, uniform slot weights, high-consequence slot-weight scaling, and an alternative exponential penalty.

\subsection{Geometric Smoothing Constant}
\label{app:geo-eps}

The AR-Geo score uses $\epsilon=10^{-5}$:
\begin{equation}
\mathrm{AR\text{-}Geo}(a)=
\exp \left(
\frac{\sum_{i\in S_a} w_i\log(\epsilon+m_i)}
{\sum_{i\in S_a} w_i}
\right).
\end{equation}
The role of $\epsilon$ is only to avoid an undefined logarithm when a component is missed.
Varying $\epsilon$ across four orders of magnitude leaves the reported scores and rankings unchanged at the precision used in the paper.
This is expected because $m_i$ is binary; once a required high-weight component is missed, the geometric score should collapse sharply regardless of the exact small constant.

\subsection{Slot Weight Perturbation}
\label{app:weight-sensitivity}

We also test whether the zero-shot ranking depends on the exact expert-derived weights in Table~\ref{tab:slot-weights}.
This matters because the weights are intentionally expert-grounded rather than learned from the evaluation set.
We compare the default weights against two alternatives.
First, Uniform-Geo assigns every non-\texttt{O} slot weight 1.0.
Second, we perturb the high-consequence slots \{callsign, runway, altitude, heading, waypoint, taxiway, condition\} by $\pm20\%$ while leaving other slots fixed.
As Table~\ref{tab:weight-sensitivity} shows, the absolute values move only slightly and the model ordering is stable.
The interpretation is that changing the coefficients changes how strongly particular misses are penalized, but it does not erase the underlying pattern: models that miss action-critical values and conditions remain weaker under consequence-aware evaluation.

\begin{table}[htbp]
\centering
\scriptsize
\begin{tabular}{lcccc}
\toprule
\textbf{Model} & \textbf{Default} & \textbf{Uniform} & \textbf{High $\times0.8$} & \textbf{High $\times1.2$} \\
\midrule
gpt-5.4        & 0.541 & 0.530 & 0.541 & 0.542 \\
\shortstack{DeepSeek-\\V4-Flash} & 0.487 & 0.473 & 0.487 & 0.488 \\
gpt-5.1        & 0.450 & 0.437 & 0.450 & 0.451 \\
claude-haiku   & 0.431 & 0.414 & 0.431 & 0.432 \\
qwen-plus      & 0.343 & 0.335 & 0.343 & 0.344 \\
gpt-4o-mini    & 0.338 & 0.318 & 0.337 & 0.340 \\
qwen3-14b      & 0.255 & 0.247 & 0.254 & 0.256 \\
qwen3-8b       & 0.109 & 0.106 & 0.108 & 0.109 \\
\bottomrule
\end{tabular}
\caption{Task~1 zero-shot AR-Geo under slot-weight perturbations. The qualitative conclusions do not depend on the exact weight vector.}
\label{tab:weight-sensitivity}
\end{table}

\subsection{Required-Slot Strictness}
\label{app:schema-sensitivity}

We also test whether the Task~1 rankings are driven by an overly strict action schema.
The default schema scores all gold slots associated with each action type, including execution conditions when they are present.
We compare it with two relaxed variants.
\textit{NoCond} makes execution conditions optional for all actions.
\textit{CoreOnly} further keeps only the core executable fields for each action, such as altitude for \texttt{altitude\_control}, heading for \texttt{heading\_control}, frequency for \texttt{contact}, and runway or route targets for runway, taxiway, and clearance actions.
These variants are not intended to replace the primary metric; they test whether the ranking is an artifact of requiring contextual fields too aggressively.

\begin{table*}[htbp]
\centering
\scriptsize
\begin{tabular}{llcccccccc}
\toprule
\textbf{Schema} & \textbf{Metric} & \textbf{gpt-5.4} & \textbf{\shortstack{DeepSeek-\\V4-Flash}} & \textbf{gpt-5.1} & \textbf{claude} & \textbf{qwen+} & \textbf{gpt-4o} & \textbf{qwen14b} & \textbf{qwen8b} \\
\midrule
Default & AR-Lin & 0.774 & 0.746 & 0.722 & 0.719 & 0.639 & 0.667 & 0.561 & 0.290 \\
Default & AR-Geo & 0.540 & 0.487 & 0.450 & 0.431 & 0.343 & 0.338 & 0.255 & 0.109 \\
NoCond & AR-Lin & 0.824 & 0.810 & 0.785 & 0.797 & 0.699 & 0.754 & 0.610 & 0.323 \\
NoCond & AR-Geo & 0.713 & 0.695 & 0.665 & 0.673 & 0.481 & 0.589 & 0.372 & 0.178 \\
CoreOnly & AR-Lin & 0.834 & 0.816 & 0.795 & 0.808 & 0.709 & 0.767 & 0.619 & 0.322 \\
CoreOnly & AR-Geo & 0.737 & 0.714 & 0.685 & 0.703 & 0.506 & 0.628 & 0.399 & 0.194 \\
\bottomrule
\end{tabular}
\caption[Task~1 zero-shot sensitivity to action-schema strictness.]{Task~1 zero-shot sensitivity to action-schema strictness. Relaxing condition and contextual slots raises absolute AR-Geo scores, but the qualitative ordering remains stable: the strongest closed/API models stay at the top and qwen3-8b remains the lowest. Only adjacent mid-ranked models swap under relaxed schemas, indicating that the main semantic-safety gap is not created by a single strict required-slot list.}
\label{tab:schema-sensitivity}
\end{table*}

\subsection{ERP as an Alternative Nonlinear Penalty}
\label{app:erp}

As a second nonlinear scoring family, we compute an Exponential Risk Penalty:
\begin{equation}
\mathrm{ERP}(a)=
\exp\left(-\lambda\sum_{i\in S_a}w_i(1-m_i)\right),
\end{equation}
with $\lambda=3.0$ for action-conditioned scores and $\lambda=2.0$ for NER scores.
ERP penalizes the total weighted miss mass, while Geo penalizes the geometric completeness of required components.
Table~\ref{tab:erp-compare} reports the zero-shot Task~1 comparison.
The rankings under Geo and ERP are consistent, so the paper uses Geo as the primary metric because it has the desired fail-safe behavior without requiring a tunable severity coefficient.

\begin{table}[htbp]
\centering
\small
\begin{tabular}{lccc}
\toprule
\textbf{Model} & \textbf{AR-Lin} & \textbf{AR-Geo} & \textbf{AR-ERP} \\
\midrule
gpt-5.4        & 0.774 & 0.541 & 0.587 \\
\shortstack{DeepSeek-\\V4-Flash} & 0.746 & 0.487 & 0.538 \\
gpt-5.1        & 0.722 & 0.451 & 0.502 \\
claude-haiku   & 0.719 & 0.431 & 0.489 \\
qwen-plus      & 0.639 & 0.343 & 0.393 \\
gpt-4o-mini    & 0.667 & 0.338 & 0.408 \\
qwen3-14b      & 0.561 & 0.255 & 0.308 \\
qwen3-8b       & 0.290 & 0.109 & 0.140 \\
\bottomrule
\end{tabular}
\caption{Task~1 zero-shot comparison of linear, geometric, and ERP action-risk scores. Geo and ERP expose the same broad semantic-safety gap, while AR-Lin is less sensitive to critical misses.}
\label{tab:erp-compare}
\end{table}

\section{Dataset Details and Examples}
\label{app:dataset-details}

The goal of the dataset is not to introduce a large general-purpose ATC benchmark.
Instead, the dataset is designed as a controlled diagnostic setting for evaluating whether models recover operationally consequential information.
We therefore prioritize safety-relevant examples, multi-action utterances, readback perturbations, and expert-auditable labels over broad coverage.
All held-out Task~1 evaluation labels are manually annotated by air traffic controllers, and Task~2 categories are validated against the same expert-grounded operational definitions.

\subsection{Annotation Protocol}
\label{app:annotation-protocol}

Task~1 annotation proceeds in two layers.
First, annotators identify the set of operational actions expressed by the utterance.
Second, they segment the utterance into contiguous chunks and assign each chunk a semantic label.
Chunk labels are assigned to the smallest continuous span that preserves the operational meaning.
For example, \textit{turn left} is annotated as a \texttt{heading\_control} action chunk, while \textit{heading two eight zero} is annotated as the \texttt{heading} value when the value is explicit.

The annotation rules are:
(i) preserve multi-action utterances rather than forcing a single dominant intent;
(ii) label callsigns even when they appear at the end of the utterance;
(iii) label numeric operational values only when explicitly present;
(iv) attach modifiers such as \textit{left}, \textit{right}, \textit{climb}, \textit{descend}, \textit{contact}, and \textit{hold short} to the action chunk when they define the action;
(v) use \texttt{condition} for execution constraints, timing, restrictions, or safety cautions;
(vi) use \texttt{relation} for connectors such as \textit{to}, \textit{via}, \textit{at}, or \textit{of} when they do not independently carry operational value;
(vii) use \texttt{attribute} for descriptors that refine an entity or procedure without being one of the core critical slots; and
(viii) use \texttt{O} for filler, greetings, discourse markers, and words outside the operational instruction.

Disfluencies such as \textit{ehm}, \textit{ah}, \textit{uh}, \textit{er}, and \textit{um} are labeled \texttt{O}.
They should not change the action set, slot labels, or risk level, matching the Task~2 \textsc{filler\_correct} category.

\subsection{Inter-Annotator Agreement}
\label{app:iaa}

Before adjudication, we audit agreement on independently annotated subsets from both tasks.
For Task~1, agreement is computed on action labels and exact slot spans; for Task~2, agreement is computed on the ordered risk level and readback error type.
Final labels are produced after controller adjudication.

\begin{table*}[t]
\centering
\small
\setlength{\tabcolsep}{5pt}
\resizebox{\textwidth}{!}{%
\begin{tabular}{llccccl}
\toprule
\textbf{Task} & \textbf{Annotation Level} & \textbf{Unit} & \textbf{Annotators} & \textbf{Metric} & \textbf{Agreement} & \textbf{Interpretation} \\
\midrule
Task~1 & Action set & Utterance & 5 ATCOs & Fleiss' $\kappa$ & 0.82 & Strong agreement \\
Task~1 & Action type & Action instance & 5 ATCOs & Macro-F1 & 0.88 & High consistency \\
Task~1 & Slot type & Token/span & 5 ATCOs & Cohen/Fleiss' $\kappa$ & 0.79 & Substantial agreement \\
Task~1 & Exact slot span & Span & 5 ATCOs & Pairwise span-F1 & 0.84 & High boundary consistency \\
Task~1 & Action-slot binding & Action-slot pair & 5 ATCOs & Pairwise F1 & 0.81 & Strong agreement \\
\midrule
Task~2 & Correctness label & Example & 5 ATCOs & Fleiss' $\kappa$ & 0.90 & Near-perfect agreement \\
Task~2 & Error type & Example & 5 ATCOs & Fleiss' $\kappa$ & 0.84 & Strong agreement \\
Task~2 & Risk level & Example & 5 ATCOs & Weighted $\kappa$ & 0.86 & Strong ordinal agreement \\
Task~2 & Affected slot & Example & 5 ATCOs & Macro-F1 & 0.82 & Strong agreement \\
\midrule
Expert weighting & Slot criticality rating & Slot type & 40 ATCOs & ICC(2,k) & 0.87 & High rating reliability \\
Expert weighting & Action criticality rating & Action type & 40 ATCOs & ICC(2,k) & 0.84 & High rating reliability \\
Expert weighting & Risk taxonomy validity & Error category & 40 ATCOs & Agreement ratio & 91.5\% & Broad expert consensus \\
\bottomrule
\end{tabular}
}
\caption{Inter-annotator agreement and expert-rating reliability for the dual-task ATC benchmark. $\kappa$ measures categorical agreement, weighted $\kappa$ accounts for ordinal risk-level distance, pairwise span-F1 evaluates boundary-sensitive span annotation, and ICC measures consistency of expert severity ratings.}
\label{tab:iaa}
\end{table*}

\subsection{Label Inventory and Ambiguous Cases}
\label{app:label-inventory}

\begin{table*}[htbp]
\centering
\small
\begin{tabular}{p{0.23\linewidth}p{0.30\linewidth}p{0.37\linewidth}}
\toprule
\textbf{Action label} & \textbf{Definition} & \textbf{Typical cues} \\
\midrule
\texttt{altitude\_control} & Assigning, requesting, reporting, or constraining altitude or flight level. & \textit{climb}, \textit{descend}, \textit{maintain}, \textit{flight level}. \\
\texttt{heading\_control} & Assigning or requesting aircraft direction, vector, turn, or heading. & \textit{turn left}, \textit{fly heading}, \textit{continue present heading}. \\
\texttt{speed\_control} & Assigning or modifying speed. & \textit{reduce speed}, \textit{maintain speed}, \textit{knots}. \\
\texttt{runway\_instruction} & Runway use, crossing, line-up, hold-short, landing, or runway vacating instructions. & \textit{line up}, \textit{hold short}, \textit{cross runway}. \\
\texttt{taxiway\_instruction} & Ground-movement routing or taxi instructions. & \textit{taxi via}, \textit{holding point}, taxiway letters. \\
\texttt{clearance} & Operational clearance or approach authorization. & \textit{cleared for}, \textit{cleared ILS}, \textit{direct}. \\
\texttt{contact} & Frequency or controller handoff. & \textit{contact tower}, \textit{monitor ground}. \\
\texttt{sequence\_control} & Traffic sequencing, following, or ordering. & \textit{number two}, \textit{follow traffic}, \textit{behind the Airbus}. \\
\bottomrule
\end{tabular}
\caption{Task~1 utterance-level action inventory. Utterances may contain multiple actions.}
\label{tab:action-inventory}
\end{table*}

\begin{table*}[htbp]
\centering
\small
\begin{tabular}{p{0.19\linewidth}p{0.24\linewidth}p{0.47\linewidth}}
\toprule
\textbf{Slot label} & \textbf{Role} & \textbf{Definition} \\
\midrule
\texttt{callsign} & target & Aircraft identifier addressed by or reporting to the controller. \\
\texttt{runway} & target & Runway identifier or runway-like landing/crossing target. \\
\texttt{taxiway} & target & Taxiway letters or ground-routing path elements. \\
\texttt{waypoint} & target & Named fix, holding point, gate, intersection, or route target. \\
\texttt{altitude} & value & Assigned, reported, or constrained altitude or flight level. \\
\texttt{heading} & value & Numeric or named aircraft direction, heading, vector, or present-heading reference. \\
\texttt{speed} & value & Speed value or explicit speed constraint. \\
\texttt{frequency} & value & Radio frequency or channel. \\
\texttt{controller} & context & Controller, sector, tower, approach, ground, or radar unit. \\
\texttt{traffic} & context & Referenced traffic or aircraft used for sequencing or caution. \\
\texttt{condition} & condition & Timing, restriction, caution, wake turbulence, ``after passing'', ``at or above'', or similar execution condition. \\
\texttt{relation} & connector & Functional connector linking action and target; not scored as an independent operational value. \\
\texttt{attribute} & context & Procedure or entity descriptor that refines another span but is not a core target/value. \\
\texttt{O} & outside & Filler, greeting, acknowledgement, discourse marker, or irrelevant text. \\
\bottomrule
\end{tabular}
\caption{Task~1 chunk-level label inventory. Criticality weights for scored slots are given in Table~\ref{tab:slot-weights}.}
\label{tab:slot-inventory}
\end{table*}

\begin{table*}[htbp]
\centering
\small
\begin{tabular}{p{0.27\linewidth}p{0.27\linewidth}p{0.36\linewidth}}
\toprule
\textbf{Ambiguous phrase} & \textbf{Preferred annotation} & \textbf{Rationale} \\
\midrule
\textit{turn left heading two eight zero} & \textit{turn left}: \texttt{heading\_control}; \textit{heading two eight zero}: \texttt{heading} & Directional verb phrase defines the action; numeric heading is the operational value. \\
\textit{one seventy to ripit} & \textit{one seventy}: \texttt{heading} or \texttt{heading\_control} only when no explicit action phrase exists; \textit{ripit}: \texttt{waypoint} & Bare numeric commands are preserved as operational direction, but boundary ambiguity is expected. \\
\textit{cleared for ILS approach runway three one} & \textit{cleared for ILS approach}: \texttt{clearance}; \textit{runway three one}: \texttt{runway} & Approach authorization and runway target are separated because a runway miss has direct operational consequence. \\
\textit{after landing hold short of runway two seven} & \textit{after landing}: \texttt{condition}; \textit{hold short}: \texttt{runway\_instruction}; \textit{runway two seven}: \texttt{runway} & Execution condition, action, and protected target are scored separately. \\
\textit{contact tower one one eight decimal three} & \textit{contact}: \texttt{contact}; \textit{tower}: \texttt{controller}; frequency phrase: \texttt{frequency} & Handoff action, unit, and value represent different failure modes. \\
\textit{ehm / ah / um} & \texttt{O} & Filler should not change the operational interpretation. \\
\bottomrule
\end{tabular}
\caption{Representative annotation decisions for ambiguous Task~1 cases.}
\label{tab:ambiguous-annotation}
\end{table*}

\subsection{Task 1: Structured Operational Understanding}

Task~1 contains 500 held-out ATC utterances.
Each example contains an utterance-level action list and chunk-level semantic labels.
The set is manually annotated by five air traffic controllers following the protocol above.

The action inventory contains eight operational actions.
The held-out set contains 169 altitude-control actions, 161 clearances, 141 heading-control actions, 71 speed-control actions, 59 contact actions, 50 runway instructions, 46 sequence-control actions, and 36 taxiway instructions.
Because utterances may contain multiple actions, these counts sum to more than 500.
Of the 500 utterances, 312 (62.4\%) are single-action and 188 (37.6\%) are multi-action, consistent with the operational pattern in which clearances and manoeuvre instructions are frequently combined in a single transmission.
The dominance of altitude control (33.8\%), clearance (32.2\%), and heading control (28.2\%) reflects the high frequency of these instruction types in en-route and approach ATC communications; their over-representation relative to taxiway instruction (7.2\%) and sequence control (9.2\%) mirrors real traffic composition rather than artificial balance.
At the chunk level, the most frequent non-action slots are callsign (462), condition (329), runway (180), altitude (180), controller (113), waypoint (105), heading (103), frequency (54), speed (48), taxiway (19), and traffic (10). Table~\ref{tab:task1-actions} reports utterance-level prevalence for slot categories, so repeated slots within the same utterance are counted once there.
Condition chunks appear 329 times across 264 of 500 utterances (52.8\%), confirming that execution constraints are a core feature of operational ATC instructions rather than an edge case---a distribution property that directly motivates their elevated role in our consequence-aware scoring.

Table~\ref{tab:task1-examples} gives representative Task~1 examples.
They are included to clarify the evaluation target, not to claim exhaustive coverage of all ATC phraseology.

\begin{table*}[htbp]
\centering
\small
\begin{tabular}{p{0.30\linewidth}p{0.22\linewidth}p{0.40\linewidth}}
\toprule
\textbf{Utterance} & \textbf{Actions} & \textbf{Selected labeled chunks} \\
\midrule
ex three twenty nine descent your discretion after landing two two left hold short of runway two seven follow that traffic
& altitude\_control; runway\_instruction; sequence\_control
& callsign=\textit{ex three twenty nine}; condition=\textit{your discretion after landing}; runway=\textit{two two left}; runway=\textit{runway two seven}; action=\textit{follow that traffic} \\
\midrule
care twenty one twelve descent your discretion follow that traffic caution the wake turbulence
& altitude\_control; sequence\_control
& callsign=\textit{care twenty one twelve}; action=\textit{descent}; condition=\textit{your discretion}; action=\textit{follow that traffic}; condition=\textit{caution the wake turbulence} \\
\midrule
all right sir out of three thousand cleared for the approach continental ten seventy two
& clearance
& O=\textit{all right sir}; condition=\textit{out of}; altitude=\textit{three thousand}; clearance=\textit{cleared for the approach}; callsign=\textit{continental ten seventy two} \\
\bottomrule
\end{tabular}
\caption[Representative Task~1 examples.]{Representative Task~1 examples. Chunk boundaries are evaluated exactly in the strict span metrics, while the consequence-aware action scores additionally condition slot recovery on the relevant action schema.}
\label{tab:task1-examples}
\end{table*}

\subsection{Task 2: Readback Safety Judgment}

Task~2 contains 1000 readback-verification examples constructed from source ATC instructions.
Each example includes a controller utterance, a correct readback, a pilot readback, an error type, a risk level, and the affected slot when applicable.
The test set is balanced by error type: 100 examples for each of ten categories.
The resulting risk distribution is 300 \textsc{correct}, 300 \textsc{high}, 300 \textsc{critical}, and 100 \textsc{extreme}.
Table~\ref{tab:task2-examples} shows representative examples.

\begin{table*}[htbp]
\centering
\scriptsize
\begin{tabular}{@{}p{0.28\linewidth}p{0.28\linewidth}p{0.16\linewidth}p{0.16\linewidth}@{}}
\toprule
\textbf{Controller instruction} & \textbf{Pilot readback} & \textbf{Error type} & \textbf{Risk level} \\
\midrule
austrian seven zero seven p descend altitude four thousand feet qnh one zero two one cleared for ils approach runway three one
& descend cleared for ils approach altitude five thousand feet runway three one qnh one zero two one austrian seven zero seven p
& VALUE\_CRITICAL & CRITICAL \\
\midrule
sabena seven eight one six turn left twenty degrees
& turn nineteen degrees sabena seven eight one six
& VALUE\_CRITICAL & CRITICAL \\
\midrule
skytravel six one eight praha radar contact climb flight level three one zero
& climb flight level four one zero radar contact skytravel six one eight
& VALUE\_CRITICAL & CRITICAL \\
\bottomrule
\end{tabular}
\caption{Representative Task~2 readback examples. The task evaluates not only whether a model detects an error, but whether it assigns the correct operational severity.}
\label{tab:task2-examples}
\end{table*}

The Task~2 fine-tuning split contains 2853 training examples and 90 validation examples.
The training set is kept separate from the balanced evaluation set and includes examples across all risk levels and error categories.
The relatively small number of condition-sensitive cases remains one reason \textsc{constraint\_high} is difficult after fine-tuning.

\subsection{Case-Level Validation Examples}
\label{app:case-validation}

Table~\ref{tab:case-validation} gives representative cases used during qualitative validation.
They illustrate why the benchmark separates strict boundary failures from value disappearance and why Task~2 evaluates directional risk calibration rather than only error detection.

\begin{table*}[htbp]
\centering
\scriptsize
\setlength{\tabcolsep}{3pt}
\begin{tabular}{p{0.18\linewidth}p{0.30\linewidth}p{0.22\linewidth}p{0.22\linewidth}}
\toprule
\textbf{Case type} & \textbf{Example pattern} & \textbf{Surface metric behavior} & \textbf{Operational interpretation} \\
\midrule
Altitude omitted & ``climb FL250'' predicted without \texttt{altitude} & Several surrounding tokens still match & Critical value disappears \\
Condition omitted & ``after ALPHA'' omitted from an otherwise correct command & High overlap on action and values & Execution precondition is lost \\
Callsign substitution & ``BAW123'' read back as ``BAW132'' & Small token-level edit & Wrong aircraft may execute instruction \\
Runway substitution & ``runway 27'' read back as ``runway 22'' & Local numeric mismatch & Critical target changes \\
Heading boundary failure & ``turn left heading 270'' span split across action and value & Strict span score drops & Operational value may still be recoverable \\
Heading disappearance & Heading action predicted but \texttt{heading=270} missing & Action type remains correct & Directional assignment is unusable \\
Frequency error & ``contact tower 118.7'' read back as ``118.1'' & Clear value mismatch & Risk is high but usually lower than altitude/runway/callsign \\
Harmless filler & ``uh, BAW123, climb FL250'' & Extra filler tokens reduce exact surface match & Operational meaning unchanged \\
Number formatting & ``flight level two five zero'' vs.\ ``FL250'' & Span form differs & Equivalent value should remain correct \\
Multi-error & Callsign and altitude both wrong in one readback & Error type may be detected & Severity should escalate to extreme \\
\bottomrule
\end{tabular}
\caption{Representative validation cases. The examples are schematic patterns rather than new test items; they summarize the case-level checks used to audit whether the metrics reflect operational consequence.}
\label{tab:case-validation}
\end{table*}

\section{Metric Definitions and Worked Example}
\label{app:metric-details}

This section gives the exact matching and aggregation rules used by the Task~1 metrics.
All span-based metrics require exact text span and label agreement after the same normalization used by the released evaluator: lower-casing, whitespace normalization, and JSON schema normalization.
Partial span overlap is not counted as a strict match.

\subsection{Span Matching and Duplicate Slots}

Let $\mathcal{G}$ be the multiset of gold labeled spans and $\mathcal{P}$ be the multiset of predicted labeled spans.
Each span is represented as a pair $(x,\ell)$, where $x$ is the normalized contiguous text span and $\ell$ is the label.
Matching is one-to-one: if the same slot type appears multiple times, each predicted span can match at most one gold span with identical text and label.
This prevents a model from receiving credit twice for predicting one runway when the utterance contains two runway references.
Unmatched gold spans count as false negatives; unmatched predicted spans count as false positives.

\subsection{NER-Lin and NER-Geo}

The linear risk-weighted NER score uses weighted precision and recall:
\begin{align}
P_{\mathrm{lin}} &=
\frac{\sum_{s\in\mathcal{M}} w(s)}
{\sum_{s\in\mathcal{P}} w(s)},\\
R_{\mathrm{lin}} &=
\frac{\sum_{s\in\mathcal{M}} w(s)}
{\sum_{s\in\mathcal{G}} w(s)},\\
\mathrm{NER\text{-}Lin} &=
\frac{2P_{\mathrm{lin}}R_{\mathrm{lin}}}{P_{\mathrm{lin}}+R_{\mathrm{lin}}},
\end{align}
where $\mathcal{M}$ is the one-to-one matched set and $w(s)$ is the slot weight.
If the denominator is zero, the corresponding precision or recall term is defined as zero unless both prediction and gold are empty, in which case the instance is treated as trivially correct for that component.

The geometric NER score is computed over gold spans as a completeness score:
\begin{equation}
\mathrm{NER\text{-}Geo} =
\exp\left(
\frac{\sum_{s\in\mathcal{G}} w(s)\log(\epsilon+m_s)}
{\sum_{s\in\mathcal{G}} w(s)}
\right),
\end{equation}
where $m_s=1$ if gold span $s$ is matched and $0$ otherwise.
We use $\epsilon=10^{-5}$.
NER-Geo is recall-oriented because the operational question is whether required safety-relevant information was recovered.

\subsection{AR-Lin, AR-Geo, and Strict}

Action-risk metrics condition slot recovery on the gold action schema.
For each gold action instance $a$, let $S_a$ be the set of gold slots associated with that action.
If action $a$ is not predicted in the utterance-level action set, both AR-Lin and AR-Geo for that action are zero.
If the action is predicted, each slot $i\in S_a$ receives $m_i=1$ when the corresponding span and label are matched and $m_i=0$ otherwise.

\begin{align}
\mathrm{AR\text{-}Lin}(a) &=
\frac{\sum_{i\in S_a} w_i m_i}
{\sum_{i\in S_a} w_i},\\
\mathrm{AR\text{-}Geo}(a) &=
\exp\left(
\frac{\sum_{i\in S_a} w_i\log(\epsilon+m_i)}
{\sum_{i\in S_a} w_i}
\right).
\end{align}
Utterance-level AR scores average over gold action instances, and corpus-level AR scores average over utterances.
Predicted extra actions do not create additional gold action instances, but they can still hurt Action-Exact and can introduce false-positive spans in NER-Lin.

Strict is the fraction of utterances for which all gold actions are predicted and every action-conditioned gold slot is matched.
It is intentionally unforgiving:
\begin{equation}
\mathrm{Strict}(u)=
\mathbf{1}\!\left[
\substack{\hat{A}_u=A_u\\
\land\ \forall a\in A_u,\forall i\in S_a,\ m_i=1}
\right].
\end{equation}
Strict is useful as a complete-recovery indicator, but it is not the primary metric because it does not distinguish near-complete recovery from catastrophic misses.

\subsection{Bootstrap Confidence Intervals}
\label{app:bootstrap-ci}

To quantify uncertainty around the main reported differences, we compute item-level nonparametric bootstrap confidence intervals with 1,000 resamples.
For Task~1, each resample draws 500 utterances with replacement and recomputes AR-Geo.
For Task~2, each resample draws readback examples with replacement and recomputes RL Acc, DDR, and WDS.
Table~\ref{tab:bootstrap-ci} reports representative intervals for the strongest zero-shot Task~1 model, a strong API baseline, the weakest zero-shot open model, and the key Task~2 prompt/fine-tuning comparisons.

\begin{table*}[htbp]
\centering
\scriptsize
\begin{tabular}{lllcc}
\toprule
\textbf{Task} & \textbf{Model / Setting} & \textbf{Metric} & \textbf{Point} & \textbf{95\% bootstrap CI} \\
\midrule
Task~1 & gpt-5.4 ZS & AR-Geo & 0.541 & [0.504, 0.578] \\
Task~1 & DeepSeek-V4-Flash ZS & AR-Geo & 0.487 & [0.445, 0.527] \\
Task~1 & qwen3-8b ZS & AR-Geo & 0.109 & [0.086, 0.133] \\
\midrule
Task~2 & DeepSeek-V4-Flash ZS & RL Acc & 0.646 & [0.604, 0.686] \\
Task~2 & DeepSeek-V4-Flash ZS & DDR & 0.100 & [0.069, 0.132] \\
Task~2 & DeepSeek-V4-Flash ZS & WDS & 0.034 & [0.024, 0.046] \\
Task~2 & DeepSeek-V4-Flash Full-Aligned & RL Acc & 0.878 & [0.848, 0.906] \\
Task~2 & DeepSeek-V4-Flash Full-Aligned & DDR & 0.057 & [0.034, 0.083] \\
Task~2 & DeepSeek-V4-Flash Full-Aligned & WDS & 0.022 & [0.013, 0.032] \\
Task~2 & qwen3-8b Fine-tuned & RL Acc & 0.906 & [0.878, 0.928] \\
Task~2 & qwen3-8b Fine-tuned & DDR & 0.083 & [0.056, 0.115] \\
Task~2 & qwen3-8b Fine-tuned & WDS & 0.029 & [0.019, 0.039] \\
\bottomrule
\end{tabular}
\caption[Bootstrap confidence intervals for key consequence-aware metrics.]{Bootstrap confidence intervals for key consequence-aware metrics. The intervals support the main ranking differences, especially the large Task~1 gap between frontier models and qwen3-8b zero-shot, and the Task~2 RL Acc improvement from zero-shot to Full-Aligned prompting or fine-tuning.}
\label{tab:bootstrap-ci}
\end{table*}

\subsection{Worked Example}

Consider the gold utterance:
\begin{quote}
\small
\textit{air malta five three nine turn right heading two four zero contact tower one one eight decimal three}
\end{quote}
The gold action set is \{\texttt{heading\_control}, \texttt{contact}\}.
The relevant gold slots are callsign ($w=1.0$), heading value ($w=0.8$), controller ($w=0.2$), and frequency ($w=0.4$).

Table~\ref{tab:worked-example} compares two model outputs.
Both make one error, but the operational consequences differ.
Prediction A misses the frequency value while recovering the heading.
Prediction B recovers the frequency but misses the heading value.

\begin{table*}[htbp]
\centering
\small
\begin{tabular}{lccc}
\toprule
\textbf{Prediction} & \textbf{Recovered weighted mass} & \textbf{AR-Lin} & \textbf{AR-Geo} \\
\midrule
A: frequency missed & $1.0+0.8+0.2$ of $2.4$ & 0.833 & 0.148 \\
B: heading missed & $1.0+0.2+0.4$ of $2.4$ & 0.667 & 0.021 \\
\bottomrule
\end{tabular}
\caption[Worked example for consequence-aware scoring.]{Worked example for consequence-aware scoring. AR-Lin decreases linearly with missed weighted mass, while AR-Geo sharply penalizes missing a required operational component. The heading miss is especially severe because it removes the trajectory-defining value.}
\label{tab:worked-example}
\end{table*}

The example illustrates the intended behavior of the geometric score.
Linear scoring still assigns moderate credit when a high-consequence component is absent.
AR-Geo instead behaves like a soft completeness check: missing any required component sharply lowers the score, and missing a high-weight component lowers it more.

\section{Task 1 Additional Error Analysis}
\label{app:task1-error-analysis}

The strict slot heatmap in the main analysis should be interpreted as an exact-boundary diagnostic rather than a direct measure of operational value recovery.
The most important case is heading.
Across zero-shot models, \texttt{heading\_control} action recall is high, while strict \texttt{heading} slot recall is much lower.
Manual inspection shows that many apparent heading-slot failures are boundary failures: models often place the whole phrase \textit{turn left heading two eight zero} inside a single \texttt{heading\_control} chunk instead of splitting \textit{turn left} as the action and \textit{heading two eight zero} as the value.

This behavior is still an annotation error under strict chunk evaluation.
However, it is not always an operational-value error.
If the numeric heading remains present inside the predicted action chunk, a downstream human or parser may still recover the trajectory value.
In contrast, if the numeric heading disappears entirely, the model has lost the operational parameter.
This distinction is why the paper emphasizes action-conditioned consequence-aware scores rather than relying only on a strict slot heatmap.

Table~\ref{tab:heading-analysis} summarizes the diagnostic finding.
Among 93 held-out examples with explicit numeric heading values, the strongest zero-shot models preserve the numeric heading somewhere in the predicted \texttt{heading} or \texttt{heading\_control} region in over 90\% of cases, even though strict heading-slot recall is much lower.

\begin{table}[htbp]
\centering
\small
\begin{tabular}{lcc}
\toprule
\textbf{Model} & \textbf{Numeric heading preserved} & \textbf{Share} \\
\midrule
\shortstack{DeepSeek-\\V4-Flash} & 87 / 93 & 0.94 \\
gpt-5.4 & 86 / 93 & 0.92 \\
claude-haiku & 86 / 93 & 0.92 \\
gpt-5.1 & 85 / 93 & 0.91 \\
qwen-plus & 84 / 93 & 0.90 \\
gpt-4o-mini & 83 / 93 & 0.89 \\
qwen3-14b & 82 / 93 & 0.88 \\
qwen3-8b & 69 / 93 & 0.74 \\
\bottomrule
\end{tabular}
\caption[Value-aware heading diagnostic on Task~1 zero-shot outputs.]{Value-aware heading diagnostic on Task~1 zero-shot outputs. Many strict heading-slot failures preserve the numeric value inside a neighboring action chunk, motivating action-conditioned analysis.}
\label{tab:heading-analysis}
\end{table}

\subsection{What Causes AR-Geo Collapse?}
\label{app:argeo-collapse}

Because AR-Geo is intentionally nonlinear, we inspect whether low scores are usually caused by one high-consequence miss or by many small misses.
We define an action instance as collapsed when its action-conditioned AR-Geo is below 0.10.
Across the eight zero-shot models, 3,330 of 5,864 gold action instances meet this threshold.
Among these collapsed instances, 2,033 (61.1\%) are caused by a single missed slot and 1,297 (38.9\%) by multiple missed slots.
Thus, AR-Geo frequently exposes a one-field operational failure rather than merely accumulating many small boundary errors.

\begin{table}[htbp]
\centering
\scriptsize
\begin{tabular}{lcc}
\toprule
\textbf{Missed slot} & \textbf{Count} & \textbf{Share} \\
\midrule
condition & 1684 & 50.6\% \\
callsign & 733 & 22.0\% \\
heading & 596 & 17.9\% \\
runway & 527 & 15.8\% \\
altitude & 512 & 15.4\% \\
waypoint & 220 & 6.6\% \\
speed & 112 & 3.4\% \\
frequency & 70 & 2.1\% \\
\bottomrule
\end{tabular}
\caption[Slot misses associated with collapsed AR-Geo action instances.]{Slot misses associated with collapsed AR-Geo action instances in the eight-model zero-shot comparison. Shares need not sum to 100\% because a collapsed action can miss multiple slots. Conditions are the most frequent source, while heading, runway, and altitude account for many single-field safety collapses.}
\label{tab:argeo-collapse}
\end{table}

\begin{table*}[htbp]
\centering
\scriptsize
\setlength{\tabcolsep}{3pt}
\begin{tabular}{@{}p{0.18\textwidth}p{0.14\textwidth}p{0.17\textwidth}cccp{0.17\textwidth}@{}}
\toprule
\textbf{Model and utterance} & \textbf{Gold slot} & \textbf{Prediction} & \textbf{NER-F1} & \textbf{AR-Lin} & \textbf{AR-Geo} & \textbf{Interpretation} \\
\midrule
qwen-plus (Full-Aligned): \textit{csa four nine three continue approach runway one three call tower one one eight one} & \texttt{controller = tower} & Frequency and action preserved; controller label omitted & 0.909 & 0.971 & 0.824 & Low-weight controller miss leaves the operational instruction largely recoverable. \\
Qwen3-8B: \textit{speedbird one two nine contact rhein on one three two decimal four} & \texttt{callsign = speedbird one two nine} & \texttt{callsign = speedbird}; number split as \texttt{attribute} & 0.727 & 0.375 & 0.0007 & Aircraft identity is incomplete. \\
Claude-Haiku-4.5: \textit{climb flight level three four zero to be level by kilnu norshuttle four five zero two} & \texttt{callsign = norshuttle four five zero two}; \texttt{waypoint = kilnu} & Callsign missing; \texttt{waypoint = kilnu norshuttle} & 0.889 & 0.600 & 0.0100 & Climb instruction becomes unattributed. \\
DeepSeek-V4-Flash: multi-instruction speed, heading, runway, and descent clearance with QNH & shared \texttt{condition = qnh one zero two three} & QNH absorbed into altitude; shared condition missing & 0.889 & 0.615 & 0.0160 & A shared safety constraint is omitted. \\
\bottomrule
\end{tabular}
\caption{Contrastive Task~1 examples for AR-Geo. Scores are computed with the same evaluator used in the main results. A low-weight controller-label miss retains a high AR-Geo score, while missing a high-weight callsign or shared execution condition causes the score to collapse despite substantial NER-F1.}
\label{tab:argeo-failure-examples}
\end{table*}

\section{Task 2 Taxonomy and Examples}
\label{app:task2-taxonomy}

Task~2 evaluates controlled readback safety judgment rather than string similarity.
The model must determine whether the pilot readback preserves the operational meaning of the controller instruction.
Each example is labeled with \texttt{is\_correct}, \texttt{error\_type}, \texttt{risk\_level}, and \texttt{affected\_slot}.
The affected slot is the most safety-relevant field whose value, target, or condition is wrong or missing.
For multi-error cases, the affected slot is set to the highest-consequence affected field when one dominates; otherwise it is recorded as multi-slot.

\begin{table*}[htbp]
\centering
\scriptsize
\begin{tabular}{@{}p{0.17\linewidth}p{0.10\linewidth}p{0.30\linewidth}p{0.31\linewidth}@{}}
\toprule
\textbf{Error type} & \textbf{Risk} & \textbf{Definition} & \textbf{Representative example} \\
\midrule
\textsc{correct} & correct & The readback preserves all operationally required fields. & Instruction: \textit{climb flight level two five zero}; readback: \textit{climb flight level two five zero}. \\
\textsc{filler\_correct} & correct & The readback adds harmless fillers or acknowledgements without changing operational content. & \textit{uh climb flight level two five zero}. \\
\textsc{number\_format} & correct & The readback changes pronunciation or formatting but preserves the same numeric value. & \textit{one two zero decimal two seven} vs.\ \textit{one twenty decimal two seven}. \\
\textsc{value\_critical} & critical & A critical value such as altitude, heading, runway, or callsign-relevant numeric value is substituted. & \textit{descend four thousand} read back as \textit{descend five thousand}. \\
\textsc{omission\_critical} & critical & A critical required field is omitted. & \textit{turn heading two seven zero} read back without the heading. \\
\textsc{target\_critical} & critical & The target aircraft, runway, waypoint, or protected target is wrong. & \textit{runway three one} read back as \textit{runway one three}. \\
\textsc{value\_high} & high & A lower-criticality value such as speed or frequency is wrong. & \textit{contact one two zero decimal two seven} read back as \textit{one two zero decimal seven two}. \\
\textsc{omission\_high} & high & A lower-criticality field is omitted. & Speed restriction or frequency omitted while core clearance remains intact. \\
\textsc{constraint\_high} & high & An execution condition, restriction, or timing constraint is dropped or changed. & \textit{after passing ALPHA climb FL250} read back as \textit{climb FL250}. \\
\textsc{multi\_error} & extreme & Two or more consequential fields are wrong or omitted simultaneously. & Wrong altitude and wrong runway in the same readback. \\
\bottomrule
\end{tabular}
\caption[Task~2 readback taxonomy.]{Task~2 readback taxonomy. \textsc{number\_format} and \textsc{filler\_correct} are benign by design; models should not over-penalize harmless surface variation. \textsc{multi\_error} is \textsc{extreme} because simultaneous critical failures reduce the chance of safe recovery.}
\label{tab:task2-taxonomy}
\end{table*}

Two benign categories are intentionally included.
\textsc{number\_format} tests whether a model can recognize equivalent spoken-number forms.
\textsc{filler\_correct} tests whether disfluency causes false alarms.
Both are operationally correct when the underlying values and targets are preserved.
At the other end, \textsc{multi\_error} is labeled \textsc{extreme} because simultaneous independent errors, such as a wrong runway and a wrong altitude, create compounded operational risk and are less likely to be recoverable through a single clarification.

\section{Fine-Tuning Details and Loss Functions}
\label{app:finetune}

\subsection{SFT Data Format}

Both fine-tuning tasks use supervised instruction tuning with the same high-level chat format.
The system message contains the task prompt, the user message contains the utterance or readback pair, and the assistant message is a single valid JSON object.
For Task~1, the assistant JSON contains the utterance, the predicted action list, and chunk-level labeled spans.
For Task~2, the assistant JSON contains \texttt{is\_correct}, \texttt{error\_type}, \texttt{risk\_level}, \texttt{affected\_slot}, and a short explanation.
Only assistant tokens are included in the loss; prompt and user tokens are masked with label value $-100$.

\begin{verbatim}
system: task prompt and output schema
user:   utterance or readback pair
assistant: valid JSON object
\end{verbatim}

\subsection{Task 1 Fine-Tuning}

Task~1 fine-tuning uses Qwen3-8B with LoRA adapters on a single NVIDIA A800 (80\,GB) GPU.
The training set contains 2,035 examples and excludes the held-out 500-example test set.
The LoRA configuration is rank 32, alpha 64, dropout 0.05, and all linear target modules.
Training uses bf16, batch size 8, gradient accumulation 2, learning rate $10^{-4}$, cosine schedule, warmup ratio 0.05, and 3 epochs (wall time $\sim$89 min for Risk-Loss; the CE baseline uses 2 epochs, $\sim$57 min).
The maximum sequence length is 8192 to accommodate the full structured prompt.

Table~\ref{tab:task1-main} reports two Task~1 fine-tuning variants.
\textit{FT (CE)} is a standard next-token cross-entropy baseline (AR-Geo 0.515).
\textit{FT (Risk-Loss)} applies the same consequence-weighted cross-entropy used in Task~2, with higher token weights on altitude, heading, callsign, runway, and waypoint spans (AR-Geo 0.686).
The risk-loss variant substantially outperforms the CE baseline (+17.1pp AR-Geo), indicating that the consequence-weighting objective transfers effectively to structured slot extraction.

\subsection{Task 2 Fine-Tuning}

Task~2 fine-tuning also uses Qwen3-8B with LoRA rank 32, alpha 64, dropout 0.05, bf16 training, batch size 8, gradient accumulation 2, learning rate $10^{-4}$, cosine schedule, and warmup ratio 0.05, on the same A800 GPU.
The training set contains 2853 examples and the validation split contains 90 examples.
The run uses 5 epochs because Task~2 is a compact classification-style JSON generation problem (wall time $\sim$17 min).

\begin{table}[htbp]
\centering
\small
\begin{tabular}{lcc}
\toprule
\textbf{Setting} & \textbf{Task 1 FT} & \textbf{Task 2 FT} \\
\midrule
Base model & Qwen3-8B & Qwen3-8B \\
Adapter & LoRA & LoRA \\
Rank / alpha & 32 / 64 & 32 / 64 \\
Dropout & 0.05 & 0.05 \\
Target modules & all linear & all linear \\
Precision & bf16 & bf16 \\
Batch size & 8 & 8 \\
Grad. accumulation & 2 & 2 \\
Learning rate & $10^{-4}$ & $10^{-4}$ \\
Schedule & cosine & cosine \\
Warmup ratio & 0.05 & 0.05 \\
Max seq. length & 8192 & 2048 \\
Epochs & 3 & 5 \\
Train / validation & 2,035 / 200 & 2,853 / 90 \\
\midrule
Hardware & \multicolumn{2}{c}{1$\times$ NVIDIA A800 (80\,GB)} \\
Wall time & $\sim$89 min & $\sim$17 min \\
\bottomrule
\end{tabular}
\caption{Fine-tuning hyperparameters. Task~2 uses more epochs because the dataset is smaller and the output space is a compact classification-style JSON schema. Wall time is measured on a single A800 GPU.}
\label{tab:ft-hparams}
\end{table}

\subsection{Next-Token Loss vs. Risk-Aware Loss}

Table~\ref{tab:task1-main} reports both a CE baseline and a risk-aware-loss variant for Task~1 fine-tuning.
When trained with the correct context length (max\_length=8192), the risk-aware variant achieves AR-Geo 0.686 versus 0.515 for the CE baseline---a gain of +17.1pp.
The largest improvement is heading recall (20.8\%$\to$67.7\%), which is directly explained by the higher token weight assigned to heading spans during training.

This result shows that the same consequence-aware structure used to design the evaluation metric can also improve training efficiency.
By concentrating gradient signal on safety-critical slots, the risk-aware objective teaches the model to attend to operational values that a standard cross-entropy loss treats equally with low-weight surface tokens.

Task~2 benefits from analogous risk-aware weighting at a different granularity.
There, the weighted fields are whole JSON keys (\texttt{risk\_level}, \texttt{error\_type}, \texttt{affected\_slot}) and sample-level severity multipliers, because the evaluation target is a classification decision rather than a token boundary.
The common principle across both tasks is the same: up-weighting the output dimensions that correspond to safety-relevant operational distinctions leads to better consequence-aware performance at test time.

\subsection{Risk-Aware Token Loss}

The risk-aware trainer replaces the standard mean cross-entropy with per-token weighted cross-entropy:
\begin{equation}
\mathcal{L}_{\mathrm{risk}} =
\frac{\sum_i \alpha_i \, \mathrm{CE}(z_i,y_i)\mathbf{1}[y_i\neq -100]}
{\sum_i \alpha_i \mathbf{1}[y_i\neq -100]}.
\end{equation}
Here $\alpha_i$ is a token-level risk weight derived from the assistant response.
For Task~1, slot text and label tokens receive weights based on operational criticality:
altitude, heading, callsign, and runway receive the highest weights;
waypoint and taxiway receive intermediate weights;
frequency, speed, traffic, controller, and \texttt{O} receive lower weights.
Action label tokens receive an action weight.

For Task~2, the weighting focuses on safety judgment fields:
\texttt{risk\_level} receives weight 2.0, \texttt{error\_type} receives 1.5, \texttt{affected\_slot} receives at least 1.5, \texttt{is\_correct} receives 1.0, and explanation text receives 0.5.
The sample-level multiplier is EXTREME $\times2.0$, CRITICAL $\times1.5$, HIGH $\times1.0$, and CORRECT $\times0.7$.
This design intentionally up-weights dangerous under-recognition cases during training.

\section{Expert Questionnaire}
\label{app:questionnaire}

We use an expert questionnaire to validate the slot weights, action criticality assumptions, readback taxonomy, and nonlinear metric design.
The survey was administered through an online questionnaire tool, and the anonymized response export is included in the supplementary materials.
The controller questionnaire was conducted on a voluntary basis for expert validation of evaluation criteria. Responses were anonymized before analysis, and no personally identifying information about respondents is released.
The export contains 40 valid responses from air traffic controllers across China (30), Singapore (5), and India (5).
The respondents cover aerodrome/tower control (17), approach/terminal control (13), area/en-route control (8), and instructor/examiner roles (2).
Experience levels range from less than two years to more than twenty years: 2 respondents report less than 2 years, 8 report 2--5 years, 14 report 6--10 years, 12 report 11--20 years, and 4 report more than 20 years.

\begin{table}[htbp]
\centering
\scriptsize
\begin{tabular}{llr}
\toprule
\textbf{Dimension} & \textbf{Category} & \textbf{N} \\
\midrule
Country & China & 30\\
Country & Singapore & 5 \\
Country & India & 5 \\
\midrule
Control background & Aerodrome / tower control & 17 \\
Control background & Approach / terminal control & 13 \\
Control background & Area / en-route control & 8 \\
Control background & ATC instructor / examiner & 2 \\
\midrule
Experience & Less than 2 years & 2 \\
Experience & 2--5 years & 8 \\
Experience & 6--10 years & 14 \\
Experience & 11--20 years & 12 \\
Experience & More than 20 years & 4 \\
\bottomrule
\end{tabular}
\caption{Background of the 40 controller questionnaire respondents.}
\label{tab:controller-background}
\end{table}

\subsection{Survey Structure}

The survey contains six parts:
\begin{itemize}
    \item \textbf{Part A: Professional background.} Role, country, and years of ATC experience.
    \item \textbf{Part B: Slot risk weights.} Controllers rate callsign, altitude, runway, waypoint, taxiway, heading, condition, frequency, and speed on a 0--10 severity scale.
    \item \textbf{Part C: Action type risk.} Controllers rate the eight action types used in the framework on a 0--10 severity scale.
    \item \textbf{Part D: Recognition error impact.} Controllers rate concrete single-error scenarios for callsign, heading, altitude, speed, and frequency errors.
    \item \textbf{Part E: Readback error taxonomy.} Controllers classify safety-critical, non-critical, constraint, and multi-error readback failures as LOW, HIGH, CRITICAL, or EXTREME.
    \item \textbf{Part F: Metric validation.} Controllers compare a linear score with a nonlinear score in a taxiway-miss scenario and rate whether high-risk slot errors should be penalized more severely.
\end{itemize}

\subsection{Aggregation Protocol}

For numeric 0--10 ratings, we compute the sample mean, median, interquartile range (IQR), and sample variance.
The means are then normalized within each question block by the highest mean in that block.
For a slot or action item $j$ with ratings $x_{1j},\ldots,x_{nj}$, we compute
\begin{equation}
\bar{x}_j = \frac{1}{n}\sum_{i=1}^n x_{ij}, \qquad
s_j^2 = \frac{1}{n-1}\sum_{i=1}^n (x_{ij}-\bar{x}_j)^2,
\end{equation}
and the normalized criticality score
\begin{equation}
\tilde{w}_j = \frac{\bar{x}_j}{\max_k \bar{x}_k}.
\end{equation}
We then discretize the normalized scores into a small number of interpretable weights rather than using many visually precise decimals.
For Task~1 slots, the final scored weights are the tiers used in Table~\ref{tab:slot-weights}: 1.0 for the top mandatory safety-critical targets and values, 0.8 for intermediate route/surface/heading fields, 0.5 for execution conditions, 0.4 for lower-risk operational values and context fields, 0.2 for controller identifiers, and 0 for \texttt{O}.
This tiering keeps the metric reproducible and avoids over-interpreting small differences in survey means.
For example, callsign, altitude, and runway have nearly identical means (8.78--8.82) and are all assigned 1.0; waypoint and taxiway have normalized means around 0.75 and are assigned the shared 0.8 tier; frequency and speed remain nonzero lower-risk value slots.
Slots not directly asked in Part B are assigned by role: relation, attribute, and traffic are low-risk context fields with weight 0.4, controller is assigned 0.2, and \texttt{O} is unscored.

\begin{figure}[htbp]
\centering
\includegraphics[width=\columnwidth]{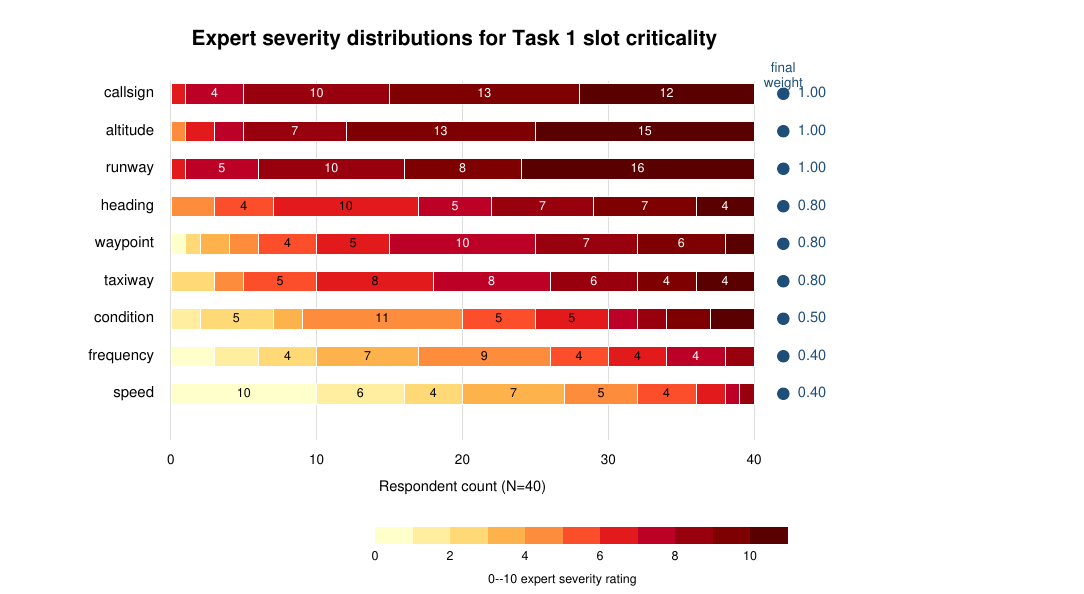}
\caption[Distribution of expert severity ratings for Task~1 slot criticality.]{Distribution of expert severity ratings for Task~1 slot criticality. The plotted responses show the empirical spread of controller ratings before normalization and tiering into final consequence weights.}
\label{fig:expert-weight-distribution}
\end{figure}

For categorical Task~2 risk labels, we report the modal label and response distribution over LOW, HIGH, CRITICAL, and EXTREME.
For the metric-validation questions, we report the mean, median, IQR, and sample variance on the 1--7 Likert scale.
Figure~\ref{fig:expert-weight-distribution} shows the distribution of raw controller ratings for Task~1 slot criticality before normalization and tiering.

\begin{table*}[htbp]
\centering
\scriptsize
\begin{tabular}{lrrrrc}
\toprule
\textbf{Task 1 slot} & \textbf{Mean} & \textbf{Median} & \textbf{IQR} & \textbf{$s^2$} & \textbf{Final w} \\
\midrule
callsign & 8.78 & 9.0 & 8.0--10.0 & 1.15 & 1.00 \\
altitude & 8.82 & 9.0 & 8.0--10.0 & 1.84 & 1.00 \\
runway & 8.82 & 9.0 & 8.0--10.0 & 1.38 & 1.00 \\
heading & 7.15 & 7.0 & 6.0--9.0 & 3.16 & 0.80 \\
waypoint & 6.65 & 7.0 & 5.5--8.0 & 4.85 & 0.80 \\
taxiway & 6.67 & 7.0 & 5.5--8.0 & 4.48 & 0.80 \\
condition & 5.10 & 4.5 & 4.0--6.5 & 6.40 & 0.50 \\
frequency & 3.90 & 4.0 & 2.5--5.5 & 4.76 & 0.40 \\
speed & 2.55 & 2.5 & 0.5--4.0 & 4.87 & 0.40 \\
\bottomrule
\end{tabular}
\caption[Part B expert ratings for Task~1 slot criticality.]{Part B expert ratings for Task~1 slot criticality. Ratings use a 0--10 severity scale. The final weights are normalized, tiered values used by the consequence-aware metrics.}
\label{tab:survey-slot-weights}
\end{table*}

\begin{table*}[htbp]
\centering
\scriptsize
\begin{tabular}{lrrrrcc}
\toprule
\textbf{Task 1 action} & \textbf{Mean} & \textbf{Median} & \textbf{IQR} & \textbf{$s^2$} & \textbf{$\tilde{w}$} & \textbf{Final w} \\
\midrule
altitude\_control & 8.65 & 9.0 & 8.0--9.0 & 0.95 & 0.98 & 1.00 \\
clearance & 8.28 & 9.0 & 6.0--10.0 & 3.85 & 0.94 & 1.00 \\
runway\_instruction & 8.85 & 9.0 & 8.0--10.0 & 1.62 & 1.00 & 1.00 \\
taxiway\_instruction & 7.22 & 7.0 & 6.0--9.0 & 5.36 & 0.82 & 0.80 \\
heading\_control & 5.40 & 5.5 & 4.0--7.0 & 4.19 & 0.61 & 0.60 \\
speed\_control & 5.35 & 5.0 & 4.0--6.0 & 5.46 & 0.60 & 0.60 \\
sequence\_control & 3.88 & 4.0 & 3.0--5.0 & 4.68 & 0.44 & 0.40 \\
contact & 2.98 & 3.0 & 1.5--4.0 & 3.77 & 0.34 & 0.30 \\
\bottomrule
\end{tabular}
\caption[Part C expert ratings for Task~1 action criticality.]{Part C expert ratings for Task~1 action criticality. Normalized scores are divided by the highest action mean before tiering.}
\label{tab:survey-action-weights}
\end{table*}

\subsection{Task 2 Risk-Level Validation}

Part E asks respondents to classify representative readback failures into ordered risk levels.
The resulting distributions support the Task~2 labels used in Table~\ref{tab:task2-types}.
Critical wrong or omitted mandatory information is most often labeled \textsc{critical}; multiple simultaneous critical errors are strongly labeled \textsc{extreme}; condition loss and non-critical value errors are most often labeled \textsc{high}, although a substantial minority of controllers choose higher categories, reflecting context dependence.
We therefore use \textsc{high} for isolated non-critical and constraint errors, while using the directional downgrade metrics in Section~\ref{sec:framework} to penalize models that under-estimate higher-consequence cases.

\begin{table*}[htbp]
\centering
\scriptsize
\begin{tabular}{llrrrr}
\toprule
\textbf{Scenario} & \textbf{Modal label} & \textbf{LOW} & \textbf{HIGH} & \textbf{CRITICAL} & \textbf{EXTREME} \\
\midrule
Critical value wrong or omitted & \textsc{critical} & 5.0 & 2.5 & 67.5 & 25.0 \\
Non-critical value wrong & \textsc{high} & 5.0 & 45.0 & 35.0 & 15.0 \\
Execution condition dropped & \textsc{high} & 5.0 & 50.0 & 37.5 & 7.5 \\
Multiple critical errors & \textsc{extreme} & 5.0 & 0.0 & 17.5 & 77.5 \\
\bottomrule
\end{tabular}
\caption[Part E expert risk-level classifications for Task~2 readback failures.]{Part E expert risk-level classifications for Task~2 readback failures. Entries are percentages over 40 respondents.}
\label{tab:survey-task2-risk}
\end{table*}

\subsection{Recognition-Error and AR-Geo Preference}

Part D presents concrete recognition-error scenarios and confirms the same severity ordering as the slot weights.
Wrong or missed altitude has the highest mean severity (8.45; median 9.0; $s^2=2.97$), followed by heading (8.20; median 8.0; $s^2=2.42$) and callsign (7.20; median 7.5; $s^2=4.68$).
Speed (3.85; median 4.0; $s^2=6.95$) and frequency (3.62; median 4.0; $s^2=4.09$) are rated lower, matching their lower Task~1 weights and their \textsc{high} rather than \textsc{critical} Task~2 treatment when they are isolated errors.

Part F directly probes whether controllers prefer a nonlinear, risk-proportional metric when a model recovers several fields but misses a high-consequence field.
On a 1--7 scale, the nonlinear-score preference has mean 5.45, median 5.0, IQR 5.0--6.5, and sample variance 1.33.
The risk-proportional-penalty question has mean 5.50, median 6.0, IQR 5.0--6.0, and sample variance 1.54.
These responses do not prove that AR-Geo is the only possible nonlinear form, but they support the design choice behind AR-Geo: correctness on lower-risk components should not fully compensate for missing a required high-risk component.
This is why AR-Geo is used as the primary Task~1 consequence-aware score, with the exponential risk penalty in Appendix~\ref{app:erp} serving as a sensitivity check.

\subsection{Controller Rating Interaction}
\label{app:controller-rating}

In addition to the questionnaire-based validation of risk weights and taxonomy design, we conduct a targeted controller-rating study for Task~1 metric alignment.
We sample 200 model predictions with stratified sampling so that the rating set covers easy, medium, and difficult cases, including high-criticality and empirically difficult slots such as heading, runway, taxiway, waypoint, and condition.
Each sampled item contains the original ATC utterance, the gold action set, the gold slot set, the model prediction, and the automatic scores used in the correlation analysis: NER-F1, rNER-F1, AR-Lin, and AR-Geo.

Controllers are shown the original instruction, the gold reference, and the model output.
They then answer the question: \textit{Does the model output preserve the operational meaning of the controller instruction?}
Ratings use a five-point operational acceptability scale, where 1 indicates that the output loses critical operational meaning and 5 indicates that the output fully preserves the instruction.
For example, a sample may show the instruction ``BAW123 climb FL250 heading 270 after ALPHA,'' the gold reference actions \texttt{altitude\_control} and \texttt{heading\_control}, and the gold slots \texttt{callsign=BAW123}, \texttt{altitude=FL250}, \texttt{heading=270}, and \texttt{condition=after ALPHA}.
If the model output only predicts \texttt{altitude\_control} with \texttt{callsign=BAW123} and \texttt{altitude=FL250}, controllers should penalize the output because it drops both the heading action/value and the execution condition.

Each item is rated independently by five controllers.
We average the raw ratings for each item and normalize the mean score into $[0,1]$:
\begin{equation}
\mathrm{human\_score} = \frac{\mathrm{mean\_rating} - 1}{4}.
\end{equation}
For instance, if a sample receives ratings 2, 2, and 3 from three available raters, the raw mean is 2.33 and the normalized score is $(2.33-1)/4=0.3325$; the same normalization is applied when all five ratings are available.
Pearson $r$ and Spearman $\rho$ are then computed between the normalized human score and each automatic metric.
The resulting correlations are reported in Table~\ref{tab:human_alignment}.

\subsection{Controller Preference Study}
\label{app:controller-preference}

We also run a paired preference study to test whether metric rankings match controller choices in cases where surface overlap and operational safety diverge.
Each item shows the same instruction and two model outputs.
Output A has higher NER-F1 but loses a safety-critical element such as altitude, runway, callsign, or execution condition; Output B has slightly lower surface overlap but preserves the operationally critical information.
Controllers choose which output is safer as an operational interpretation of the instruction.

\begin{table}[htbp]
\centering
\scriptsize
\begin{tabular}{lc}
\toprule
\textbf{Automatic ranking rule} & \textbf{Agreement with controller preference} \\
\midrule
NER-F1 & 0.41 \\
rNER-F1 & 0.58 \\
AR-Lin & 0.64 \\
AR-Geo & 0.76 \\
\bottomrule
\end{tabular}
\caption{Paired controller-preference agreement. Agreement measures whether an automatic metric ranks the controller-preferred output higher in the paired comparison.}
\label{tab:controller-preference}
\end{table}

For example, for ``BAW123 climb FL250 heading 270 after ALPHA,'' a high-overlap output that preserves \texttt{callsign=BAW123}, \texttt{heading=270}, and \texttt{waypoint=ALPHA} but drops \texttt{altitude=FL250} and the execution condition is usually rejected by controllers.
An output with a less exact boundary around the action phrase but with \texttt{altitude=FL250}, \texttt{heading=270}, and \texttt{condition=after ALPHA} preserved is preferred.
This study supports the same conclusion as the scalar rating study: operational acceptability is better explained by preserving critical units than by surface span overlap alone.

\section{Additional Task 2 Results}
\label{app:task2-detail}

Table~\ref{tab:task2-ablation} reports the Task~2 prompt ablations averaged over four models.
Full-Aligned prompting gives the best error-type F1 and the lowest downgrade rates, while the Knowledge prompt gives a comparable risk-level accuracy.

\begin{table}[htbp]
\centering
\scriptsize
\setlength{\tabcolsep}{2.2pt}
\begin{tabular}{lcccc}
\toprule
\textbf{Cond.} & \textbf{ET F1} & \textbf{RL Acc} & \textbf{DDR}$\downarrow$ & \textbf{WDS}$\downarrow$ \\
\midrule
A (ZS)      & 0.456 & 0.511 & 0.191 & 0.066 \\
C (Know.)   & 0.538 & \textbf{0.648} & 0.086 & 0.033 \\
D (Few)     & 0.525 & 0.573 & 0.125 & 0.043 \\
D+CoT       & 0.516 & 0.559 & 0.143 & 0.050 \\
B (Full)    & \textbf{0.558} & 0.646 & \textbf{0.054} & \textbf{0.019} \\
\midrule
A$\to$B     & +10.2pp & +13.5pp & $-$13.7pp & $-$4.7pp \\
\bottomrule
\end{tabular}
\caption{Task~2 prompt ablations averaged over four models. DDR/WDS are lower-is-better downgrades.}
\label{tab:task2-ablation}
\label{tab:task2-error-ablation}
\end{table}

Table~\ref{tab:task2-detail} reports per-error-type F1 for representative zero-shot models.
\textsc{constraint\_high} is consistently difficult, matching the condition-slot weakness observed in Task~1.
\textsc{multi\_error} is more reliably detected by stronger models because it often contains multiple explicit mismatches.

\begin{table}[htbp]
\centering
\scriptsize
\begin{tabular}{lccc}
\toprule
\textbf{Error Type} & \textbf{gpt-4o} & \textbf{\shortstack{DeepSeek-\\V4-Flash}} & \textbf{qwen+} \\
\midrule
\textsc{value\_critical}    & 0.226 & 0.738 & 0.560 \\
\textsc{omission\_critical} & 0.580 & 0.700 & 0.640 \\
\textsc{target\_critical}   & 0.620 & 0.760 & 0.780 \\
\textsc{value\_high}        & 0.580 & 0.620 & 0.440 \\
\textsc{omission\_high}     & 0.600 & 0.540 & 0.440 \\
\textsc{constraint\_high}   & 0.080 & 0.380 & 0.200 \\
\textsc{multi\_error}       & 0.380 & 0.920 & 0.860 \\
\textsc{filler\_correct}    & 0.260 & 0.400 & 0.300 \\
\textsc{number\_format}     & 0.480 & 0.520 & 0.320 \\
\textsc{correct}            & 0.360 & 0.260 & 0.380 \\
\bottomrule
\end{tabular}
\caption[Task~2 per-error-type F1 for representative zero-shot models.]{Task~2 per-error-type F1 for representative zero-shot models. gpt-4o = gpt-4o-mini; qwen+ = qwen-plus.}
\label{tab:task2-detail}
\end{table}

\section{Domain Transferability}
\label{app:transfer}

The consequence-aware evaluation framework is domain-open.
Adapting it to a new domain requires defining a domain action inventory, mapping domain slots to abstract criticality roles, assigning relative risk weights, and aligning an error taxonomy with domain failure modes.
The scoring formula itself is unchanged.

For medical instruction understanding, high-weight slots could include medication name, dosage, allergy status, body location, and symptom severity.
For vehicle-command or autonomous-driving communication, high-weight slots could include maneuver type, lane or spatial target, distance, obstacle reference, and temporal condition.
In each case, the key requirement is not that the ATC schema transfer directly, but that domain experts identify which pieces of language determine operational consequence.

Table~\ref{tab:transfer-recipe} gives a concrete illustration using driving-command text in the style of public Talk2Car examples only as raw language input. We do not report cross-domain model scores or claim a second validated benchmark; the example is a recipe for what must be redefined and expert-validated before consequence-aware evaluation can be used in another domain.

\begin{table*}[htbp]
\centering
\scriptsize
\begin{tabular}{@{}p{0.18\textwidth}p{0.58\textwidth}p{0.18\textwidth}@{}}
\toprule
\textbf{Pipeline step} & \textbf{Driving-command illustration} & \textbf{Purpose} \\
\midrule
Define actions & \texttt{stop}, \texttt{slow\_down}, \texttt{turn}, \texttt{change\_lane}, \texttt{park} & Action inventory \\
Map roles & TARGET = vehicle or pedestrian; VALUE = lane or speed; CONDITION = red light or safety distance & Role schema \\
Instantiate & \textit{the guy on our left is going to cross the red light, brake} $\rightarrow$ \texttt{stop}; TARGET = road user; CONDITION = red-light risk & Action units \\
Validate & Domain-expert severity elicitation and human-alignment checks & Validation gate \\
\bottomrule
\end{tabular}
\caption{Transfer recipe for adapting consequence-aware evaluation beyond ATC. The driving-command example is illustrative only and is not used as a second benchmark or scoring experiment.}
\label{tab:transfer-recipe}
\end{table*}

\section{Prompt Templates and Inference Protocol}
\label{app:prompts}

All prompting experiments use a JSON-only output contract.
Table~\ref{tab:prompt-inventory} lists the inference prompt conditions used in the ablation experiments.
Only inference prompts are documented here.
Released benchmark splits, prompts, evaluation scripts, annotation schemas, fine-tuning code, and LoRA adapters are available at \url{https://github.com/EthanChangCC/beyond-semantic-accuracy}.

\begin{table*}[htbp]
\centering
\scriptsize
\begin{tabular}{p{0.25\linewidth}p{0.13\linewidth}p{0.52\linewidth}}
\toprule
\textbf{Template} & \textbf{Condition} & \textbf{Purpose} \\
\midrule
Zero-shot & Task~1 A & Defines the action inventory, slot labels, chunking rules, and JSON output schema without demonstrations. \\
Knowledge & Task~1 C & Adds explicit domain rules, label definitions, and boundary constraints, but no worked examples. \\
Few-shot & Task~1 D & Provides representative demonstrations for ATC chunking and action extraction with minimal extra rules. \\
Full-Aligned & Task~1 B & Combines the rule block and demonstrations; used as the strongest prompt-engineered Task~1 condition. \\
Zero-shot & Task~2 A & Defines the readback verification schema and error taxonomy without examples. \\
Knowledge & Task~2 C & Adds explicit risk-level definitions and ATC safety rules for readback judgment. \\
Few-shot & Task~2 D & Adds demonstrations covering correct readbacks and each error family. \\
Few-shot+CoT & Task~2 D+CoT & Adds concise reasoning instructions before returning the final JSON object. \\
Full-Aligned & Task~2 B & Combines taxonomy rules, risk calibration guidance, and examples. \\
\bottomrule
\end{tabular}
\caption{Prompt inventory. Template names identify the prompting condition and task role used in the experiments.}
\label{tab:prompt-inventory}
\end{table*}

\subsection{Task 1 Output Contract}

Task~1 prompts require the model to output one JSON object:
\begin{verbatim}
{
  "utterance": "...",
  "utterance_action": ["..."],
  "tokens": [
    {"text": "...", "label": "callsign"},
    {"text": "...", "label": "action"},
    {"text": "...", "label": "altitude"}
  ]
}
\end{verbatim}
The allowed action labels are \texttt{altitude\_control}, \texttt{clearance}, \texttt{heading\_control}, \texttt{runway\_instruction}, \texttt{speed\_control}, \texttt{taxiway\_instruction}, \texttt{contact}, and \texttt{sequence\_control}.
The Task~1 evaluation schema includes the slot labels \texttt{callsign}, \texttt{runway}, \texttt{taxiway}, \texttt{waypoint}, \texttt{controller}, \texttt{altitude}, \texttt{speed}, \texttt{heading}, \texttt{frequency}, \texttt{traffic}, \texttt{condition}, \texttt{relation}, \texttt{attribute}, and \texttt{O}. Some fine-tuning prompts map purely functional relation/attribute tokens to \texttt{O}; evaluation uses the same released normalization and scorer for all systems.

\subsection{Task 2 Output Contract}

Task~2 prompts require the model to return:
\begin{verbatim}
{
  "is_correct": false,
  "error_type": "VALUE_CRITICAL",
  "risk_level": "CRITICAL",
  "affected_slot": "altitude",
  "explanation": "..."
}
\end{verbatim}
The error taxonomy contains ten types:
\texttt{CORRECT}, \texttt{FILLER\_CORRECT}, \texttt{NUMBER\_FORMAT}, \texttt{VALUE\_CRITICAL}, \texttt{OMISSION\_CRITICAL}, \texttt{TARGET\_CRITICAL}, \texttt{VALUE\_HIGH}, \texttt{OMISSION\_HIGH}, \texttt{CONSTRAINT\_HIGH}, and \texttt{MULTI\_ERROR}.
The risk levels are \textsc{correct}, \textsc{high}, \textsc{critical}, and \textsc{extreme}.
For all conditions, the final answer must be valid JSON and must not include free-form text outside the object.

\newread\promptpathcheck
\newcommand{\InferencePromptBlock}[3]{%
\subsection{#1}
\begin{tcolorbox}[
breakable,
enhanced,
left=1.5mm,
right=1.5mm,
top=1mm,
bottom=1mm,
boxrule=0.2mm,
colback=white,
colframe=black,
title={#2},
fonttitle=\bfseries\scriptsize
]
\openin\promptpathcheck=#3\relax
\ifeof\promptpathcheck
\closein\promptpathcheck
\VerbatimInput[
fontsize=\tiny,
breaklines=true,
breakanywhere=true,
tabsize=2
]{../#3}%
\else
\closein\promptpathcheck
\VerbatimInput[
fontsize=\tiny,
breaklines=true,
breakanywhere=true,
tabsize=2
]{#3}%
\fi
\end{tcolorbox}
}

\InferencePromptBlock
{Task 1 Zero-Shot Prompt}
{Task 1 Zero-Shot Inference Prompt}
{prompts/prompt_operational_zeroshot.txt}

\InferencePromptBlock
{Task 1 Knowledge Prompt}
{Task 1 Knowledge-Augmented Inference Prompt}
{prompts/prompt_operational_knowledge_pure.txt}

\InferencePromptBlock
{Task 1 Few-Shot Prompt}
{Task 1 Few-Shot Inference Prompt}
{prompts/prompt_operational_fewshot_pure.txt}

\InferencePromptBlock
{Task 1 Full-Aligned Prompt}
{Task 1 Full-Aligned Inference Prompt}
{prompts/prompt_operational_full_aligned.txt}

\InferencePromptBlock
{Task 2 Zero-Shot Prompt}
{Task 2 Zero-Shot Inference Prompt}
{prompts/prompt_readback_zeroshot.txt}

\InferencePromptBlock
{Task 2 Knowledge Prompt}
{Task 2 Knowledge-Augmented Inference Prompt}
{prompts/prompt_readback_knowledge.txt}

\InferencePromptBlock
{Task 2 Few-Shot Prompt}
{Task 2 Few-Shot Inference Prompt}
{prompts/prompt_readback_fewshot.txt}

\InferencePromptBlock
{Task 2 Few-Shot CoT Prompt}
{Task 2 Few-Shot CoT Inference Prompt}
{prompts/prompt_readback_fewshot_cot.txt}

\InferencePromptBlock
{Task 2 Full-Aligned Prompt}
{Task 2 Full-Aligned Inference Prompt}
{prompts/prompt_readback_full_aligned.txt}